%% file: neurips_2023.tex
\documentclass{article}

\PassOptionsToPackage{numbers, compress}{natbib}

\usepackage[final]{neurips_2023}

\usepackage[utf8]{inputenc} 
\usepackage[T1]{fontenc}    
\usepackage[hidelinks]{hyperref}       
\usepackage{url}            
\usepackage{booktabs}       
\usepackage{amsfonts}       
\usepackage{nicefrac}       
\usepackage{microtype}      
\usepackage{xcolor}         

\input{preamble.tex}

\title{The Rank-Reduced Kalman Filter: Approximate Dynamical-Low-Rank Filtering In High Dimensions}

\author{%
  Jonathan Schmidt\\ 
  Tübingen AI Center, University of Tübingen\\
  \texttt{jonathan.schmidt@uni-tuebingen.de}
  \And
  Philipp Hennig\\
  Tübingen AI Center, University of Tübingen \\
  \texttt{philipp.hennig@uni-tuebingen.de}
  \And
  Jörg Nick\\
  University of Tübingen \\
  \texttt{nick@na.uni-tuebingen.de}
  \And
  Filip Tronarp \\
  Lund University\\
  \texttt{filip.tronarp@matstat.lu.se}
}

\begin{document}

\maketitle

\input{parts/abstract.tex}

\input{parts/section_1.tex}
\input{parts/section_2.tex}
\input{parts/section_3.tex}
\input{parts/section_4.tex}

\input{parts/section_5.tex}
\input{parts/conclusion.tex}

\begin{ack}
  The authors gratefully acknowledge financial support by the German Federal Ministry of Education and Research (BMBF) through Project ADIMEM (FKZ 01IS18052B), and financial support by the European Research Council through ERC StG Action 757275 / PANAMA; the DFG Cluster of Excellence “Machine Learning - New Perspectives for Science”, EXC 2064/1, project number 390727645; the German Federal Ministry of Education and Research (BMBF) through the Tübingen AI Center (FKZ: 01IS18039A); and funds from the Ministry of Science, Research and Arts of the State of Baden-Württemberg.
  Jörg Nick is funded by the Deutsche Forschungsgemeinschaft (DFG, German Research Foundation) -- Project-ID 258734477 -- SFB 1173.
  Filip Tronarp was partially supported by the Wallenberg AI, Autonomous Systems and Software Program (WASP) funded by the Knut and Alice Wallenberg Foundation.
  The authors thank the International Max Planck Research School for Intelligent Systems (IMPRS-IS) for supporting Jonathan Schmidt.
  The authors also thank Nathanael Bosch and Christian Lubich for many valuable discussions and for helpful feedback.
\end{ack}

\bibliographystyle{apalike}
\bibliography{refs}

\cleardoublepage

\input{supplement.tex}




\end{document}

%% file: preamble.tex
\usepackage[pdftex]{graphicx}
\usepackage{amsmath}
\usepackage{amsthm}
\usepackage{amssymb}
\usepackage{mathtools}
\usepackage{mathrsfs}
\usepackage{commath}
\usepackage[capitalize]{cleveref}
\usepackage{algorithm}
\usepackage{algorithmic}
\usepackage{wrapfig}
\usepackage{subcaption}
\usepackage{multirow}
\usepackage{enumitem}

\newcommand{\q}{\quad}
\newcommand{\qq}{\qquad}

\renewcommand{\Re}{\ensuremath{\mathbb{R}}}

\newcommand{\numT}{\ensuremath{N}}

\newcommand{\N}{\ensuremath{\mathcal{N}}}

\newcommand{\mat}[1]{\ensuremath{\mathrm{\uppercase{#1}}}}
\renewcommand{\vec}[1]{\ensuremath{\mathrm{\lowercase{#1}}}}

\newcommand{\transposed}{\ensuremath{^\ast}}

\newcommand{\transposedsqrt}{\ensuremath{^{\nicefrac{\ast}{2}}}}
\newcommand{\matsqrt}{\ensuremath{^{\nicefrac{1}{2}}}}
\newcommand{\matsqrtinv}{\ensuremath{^{\nicefrac{-1}{2}}}}
\newcommand{\matsqrtinvtransposed}{\ensuremath{^{\nicefrac{-\ast}{2}}}}
\newcommand{\matinv}{\ensuremath{^{-1}}}
\newcommand{\matpseudoinv}{\ensuremath{^{+}}}

\newcommand{\statedim}{n}
\newcommand{\measdim}{m}
\newcommand{\lowrankdim}{r}
\newcommand{\diffusiondim}{q}
\newcommand{\driftmat}{\mat{A}}
\newcommand{\dispmat}{\mat{B}}

\newcommand{\transitionmat}{\ensuremath{\mat{\Phi}}}
\newcommand{\measurementmat}{\ensuremath{\mat{C}}}
\newcommand{\processnoisecov}{\ensuremath{\mat{Q}}}
\newcommand{\measurementnoisecov}{\ensuremath{\mat{R}}}
\newcommand{\predcov}{\ensuremath{\mat{\Pi}}}
\newcommand{\filtcov}{\ensuremath{\mat{\Sigma}}}
\newcommand{\lowrank}[1]{\ensuremath{#1}}
\newcommand{\australiastatedim}{30\,911}

\newtheorem{corollary}{Corollary}
\newtheorem{proposition}{Proposition}

\newtheorem{remark}{Remark}
\newtheorem{assumption}[remark]{Assumption}
\crefname{assumption}{Assumption}{Assumptions}

\newcommand{\appendixtitle}{
	\vbox{
		{\hrule height 4pt \vskip 0.25in \vskip -\parskip}
		\centering
		{\LARGE\bf The Rank-Reduced Kalman Filter: Approximate Dynamical-Low-Rank Filtering In High Dimensions\par}
		{\Large\bf Appendix\par}
		{\vskip 0.29in \vskip -\parskip \hrule height 1pt \vskip 0.09in}
	}
}

\usepackage{etoolbox}
\usepackage{pgfplots}
\usepgfplotslibrary{groupplots}
\usepackage{tikz}
\usetikzlibrary{tikzmark}
\usetikzlibrary{calc}
\definecolor{kfcolor}{rgb}{0.69,0.478,0.631}
\newcommand{\kfmarker}[2][0.75ex]{\tikz[baseline=-0.01ex]\filldraw[black,fill=kfcolor, thick,fill opacity=1.0] (0.2,0.1) -- (0.1,0)  -- (0,0.1) -- (0.1,0.2) -- cycle;}%

%% file: parts/abstract.tex
\begin{abstract}
Inference and simulation in the context of high-dimensional dynamical systems remain computationally challenging problems.
Some form of dimensionality reduction is required to make the problem tractable in general.
In this paper, we propose a novel approximate Gaussian filtering and smoothing method
which propagates low-rank approximations of the covariance matrices.
This is accomplished by projecting the Lyapunov equations associated with the prediction step to a manifold of low-rank matrices,
which are then solved by a recently developed, numerically stable, dynamical low-rank integrator.
Meanwhile, the update steps are made tractable by noting that the covariance update only transforms the column space of the covariance matrix, which is low-rank by construction.
The algorithm differentiates itself from existing ensemble-based approaches in that
the low-rank approximations of the covariance matrices are deterministic, rather than stochastic.
Crucially, this enables the method to reproduce the exact Kalman filter as the low-rank dimension approaches the true dimensionality of the problem.
Our method reduces computational complexity from cubic (for the Kalman filter) to \emph{quadratic} in the state-space size in the worst-case, and can achieve \emph{linear} complexity if the state-space model satisfies certain criteria.
Through a set of experiments in classical data-assimilation and spatio-temporal regression, we show that the proposed method consistently outperforms the ensemble-based methods in terms of error in the mean and covariance with respect to the exact Kalman filter. This comes at no additional cost in terms of asymptotic computational complexity.
\end{abstract}

%% file: parts/section_1.tex
\section{Introduction}
Spatio-temporal dynamical systems have always played an important role in the applied sciences, such as climate science, numerical weather prediction, or geophysics.
In precisely these settings, one is often faced with massive amounts of data to be processed.
At the same time the interactions are of such complex nature that it is imperative to include a notion of uncertainty in the model
and in its outputs.
Both these requirements have a reputation as being computationally expensive.
Hence, sensible approximations are indispensable at a certain scale.

Concretely, we consider state-space models of the form
\begin{subequations}\label{eq:pomp}
\begin{align}\label{eq:pomp_dynamics}
\dif \vec{x}(t) &= \driftmat \vec{x}(t) \dif t  + \dispmat \dif \vec{w}(t), \\\label{eq:pomp_measurements}
\vec{y}(t_l) \mid \vec{x}(t_l) &\sim \mathcal{N}(\measurementmat \vec{x}(t_l), \measurementnoisecov),\q l = 1, \dots, \numT
\end{align}
\end{subequations}
where $\vec{x} \in \Re^\statedim$ is the latent state, $\vec{y}\in \Re^\measdim$ is the measurement process, and $\vec{w}$ is a standard Wiener process in $\Re^{\diffusiondim}$. 
\Cref{eq:pomp_dynamics} defines the prior dynamics model via a linear time-invariant (LTI) stochastic differential equation (SDE) with drift matrix $\driftmat \in \Re^{\statedim \times \statedim}$ and dispersion matrix $\dispmat \in \Re^{\statedim \times \diffusiondim}$.
The linear Gaussian observation model \cref{eq:pomp_measurements} is specified via the measurement matrix $\measurementmat \in \Re^{\measdim \times \statedim}$ and measurement-noise covariance $\measurementnoisecov \in \Re^{\measdim \times \measdim}$, where possible time-dependency has been omitted from the notation.
While the probabilistic state-estimation problem may in principle be solved in closed form using the Kalman filter (KF) and smoother \citep{Kalman1960,Sarkka2013},
we will consider the case where the state dimension $\statedim$ and measurement dimension $\measdim$ are very large,
making the covariance recursion computationally prohibitive in practice.
Such settings include spatio-temporal Gaussian process regression \citep{Sarkka2012,Sarkka2013b,Todescato2020,Carron2016,Hamelijnck2021},
and data assimilation applications such as meteorology and oceanography \citep{Ghil1991,Evensen1994},
numerical weather forecasting \citep{Hamill2000}, geoscience \citep{Carrassi2018}, inverse problems \citep{Chada2020, Stuart2010},
and brain imaging \citep{Galka2004}.
While the Gaussian process (GP) literature offers many ways of tractable, approximate inference, approaches based on approximations of the underlying kernel function are not amenable to the present problem setting.
Without a kernel to approximate, the problem is often solved by ensemble methods,
which stochastically propagate low-rank approximations to the covariances in the filtering recursion.
This stochasticity introduces unfavorable properties in these methods.
To avert this shortcoming, we propose a novel low-rank filtering recursion, which is fully deterministic.
%
\begin{figure}
    \centering
    \includegraphics[width=.98\linewidth]{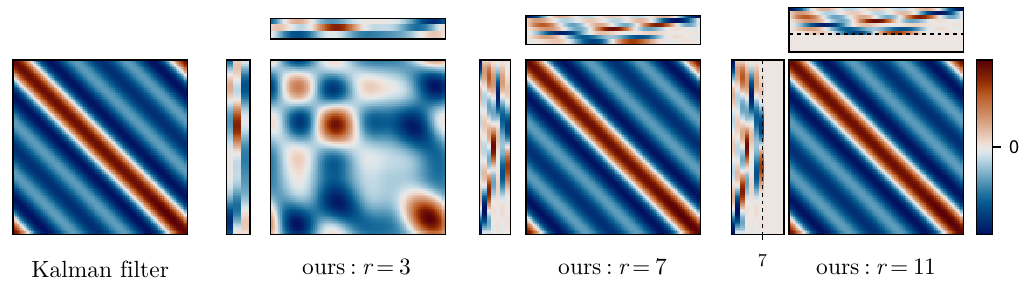}
    \caption{
    \emph{Rank-$\lowrankdim$ approximations to the true KF covariance (left) for increasing $\lowrankdim$.}
    The considered problem's true rank is $\lowrankdim^\ast = 7$ by construction, the state-dimension is $\statedim = 1000$.
    The respective low-rank factors are shown above and left of their outer product.
    For $\lowrankdim \geq \lowrankdim^\ast$, the KF estimate is recovered.
    For $\lowrankdim > \lowrankdim^\ast$, the excess columns of the low-rank factor collapse (rightmost plot).
    }
    \label{fig:fig1}
\end{figure}

\paragraph{Contributions}
This text develops a method for efficient, approximate Gaussian filtering,
including smoothing and marginal likelihood computation.
The method employs the standard predict/correct formulation of the filtering problem, and
can in principle be thought of as a square-root implementation of a Kalman filter,
in that it represents all covariance matrices as their left square-root factors \citep{Grewal2011}. 
However, instead of square matrix square roots, the covariances are approximated by $(\statedim \times\lowrankdim)$-dimensional low-rank factors, where
$\lowrankdim \ll \statedim$.
We propose an approximate low-rank prediction-correction loop, together with the computation of the backwards Markov process representing the smoothing posterior, all while preserving this low-rank structure throughout time.

Crucial to this work are (i) the use of dynamic-low-rank approximation (DLRA) \citep{Koch2007,Lubich2014,Ceruti2022} to solve Lyapunov equations on the manifold of rank-$\lowrankdim$ matrices in the prediction step
and (ii) an efficient correction scheme that updates only the low-dimensional column space of the predicted covariance factor,
which introduces no additional approximation error.
The computational complexity of the proposed filter is asymptotically equal to those of the widely-used ensemble Kalman filters.

The method offers a deterministic alternative to existing ensemble methods. 
In particular, it recovers the exact Kalman filter estimate both in the full-rank limit $\lowrankdim = \statedim$ and if the problem is in fact of rank $\lowrankdim < n$.
The latter is demonstrated in \cref{fig:fig1}, in which a problem with $\statedim = 1000$ state dimensions and a true rank of 7 is approximated with $\lowrankdim = 3, 7, 11$.
As soon as $\lowrankdim$ exceeds the true rank of the problem, the residual subspace dimensions collapse.
The method will be detailed in \cref{sec:method} and evaluated on the basis of various experiments in \cref{sec:experiments}. All source code is publicly available on GitHub.\footnote{\url{https://github.com/schmidtjonathan/RRKF.jl}}\textsuperscript{,}\footnote{\url{https://github.com/schmidtjonathan/RRKF_experiments/}}

\paragraph{Related work}\label{subsec:related-work}
The state estimation problem defined by \cref{eq:pomp} amounts to linear Gaussian state estimation and
can be computed in closed form by Kalman filtering and smoothing. A modern overview can be found in \citet{Sarkka2019}.
While the exact filtering and smoothing recursions scale linearly with the number of time points $\numT$,
they scale cubically in the state dimension $\statedim$ and the measurement dimension $\measdim$,
making their use computationally infeasible in many high-dimensional problems.
A widely-used approximate filtering technique used to circumvent this computational burden are ensemble Kalman filters (EnKF) \citep{Evensen1994, Evensen2009b, Evensen2009, Katzfuss2016, Roth2017, Carrassi2018}.
The ensemble is a set of $\lowrankdim$ randomly sampled state vectors of dimension $\statedim$, which parametrize the mean and covariance via sample statistics.
It is processed successively in a prediction-correction loop similar to the KF.
The mean-centered ensemble can be regarded as a low-rank square-root factor of the sample covariance matrix. 
In \cref{sec:experiments} we compare our method to the standard version of the EnKF and
the ensemble transform Kalman filter \citep{Bishop2001}, which is categorized as an ensemble-square-root filter \citep{Tippett2003}.

In so far as it aims for computationally tractable Gaussian process regression, the proposed method is related to
approaches using the Nystr\"om method \citep{Williams2001},
inducing point methods \citep{Snelson2005,Titsias2009,Hensman2013},
random Fourier features \citep{Rahimi2007},
and doubly-sparse variational Gaussian processes \citep{Adam2020}.

The literature offers more contributions related to the present work under the broader context of low-rank and dimensionality-reduction methods.
\citet{loiseau2018sparse,netto2018robust,vijayshankar2020dynamic} approach low-rank modelling of dynamical systems based on measured data, whereas 
\citet{farrell2001state} are concerned with low-rank approximations of a given model based on classical theory on linear time-invariant systems.
\citet{Cressie2010} develop a tractable hierarchical Bayesian model to approximate high-dimensional spatio-temporal dynamics in a reduced latent space and condition on measurements by transporting the low-dimensional diffusive process to the full space by
a projection that is assumed to be known.
In those cases, once a low rank model is obtained for the phenomena under study, state estimation becomes computationally tractable, due to the reduced size of the resulting model.
These works build upon the assumption that the state of the system evolves approximately in a low-dimensional manifold, whereas our method works under the assumption that the \emph{error} in the state estimate evolves approximately in a low-dimensional manifold.

%% file: parts/section_2.tex
\section{Background}
Throughout this work, for a function $x(t)$ and a grid $(t_1, .., t_\numT)$ we use the abbreviated notation $x(t_l) = x_l$ and $(x(t_a), ... x(t_b)) = x_{a:b}$.
The task of interest is computing the filtering densities \(p(x_l \mid y_{1:l})\) and smoothing densities \(p(x_l \mid y_{1:\numT})\)
for $l = 1, \dots, \numT$.
LTI SDEs of the form in \cref{eq:pomp_dynamics} admit an equivalent discrete formulation in terms of linear Gaussian transition densities of the form
\begin{equation}\label{eq:discr-dynamics}
    \vec{x}_l \mid \vec{x}_{l-1} \sim \N(\transitionmat_l \vec{x}_{l-1}, \processnoisecov_l).
\end{equation}
The transition matrix $\transitionmat(t) \in \Re^{\statedim \times \statedim}$ and process-noise covariance matrix $\processnoisecov(t) \in \Re^{\statedim \times \statedim}$ solve the matrix differential equations $\dot\transitionmat(t) = \driftmat\transitionmat(t)$ and
\begin{equation}\label{eq:process-noise-cov-diffeq}
    \dot\processnoisecov(t) = \driftmat\processnoisecov(t) + \processnoisecov(t)\driftmat\transposed + \dispmat\dispmat\transposed,
\end{equation}
in the interval $[t_{l-1}, t_l)$
with initial conditions \(\transitionmat(t_{l-1}) = \mathrm{I}\) and \(\processnoisecov(t_{l-1}) = \mat{0}\), respectively  \citep{Sarkka2019}.
\Cref{eq:process-noise-cov-diffeq} is known as a \emph{Lyapunov equation}, and will be of special importance in this work.

\subsection{Square-root filtering}\label{subsec:background-sqrt-kf}
For numerically stable filtering and smoothing, all covariance matrices involved can be
represented by matrix square roots (e.g. Cholesky factors).
This is known as square-root filtering/smoothing \citep{Grewal2011}.
Let $(\mat{M}_1 \q \mat{M}_2)$ denote a wide block matrix in $\Re^{d_1 \times (d_2 + d_3)}$, where $\mat{M}_1 \in \Re^{d_1 \times d_2}$ and $\mat{M}_2 \in \Re^{d_1 \times d_3}$ are some matrices. Given the filtering covariance at time $t_{l-1}$ as $\filtcov_{l-1} = \filtcov\matsqrt_{l-1}\filtcov\transposedsqrt_{l-1}$
a square-root factorization of the predicted covariance $\predcov_l = \predcov\matsqrt_l\predcov\transposedsqrt_l$ is given by
\begin{equation}
    \predcov_l = \transitionmat_l\filtcov_{l-1}\transitionmat_l\transposed + \processnoisecov_l
    = \begin{pmatrix}
        \transitionmat_l\filtcov_{l-1}\matsqrt & \processnoisecov_l\matsqrt
    \end{pmatrix}\begin{pmatrix}
        \transitionmat_l\filtcov_{l-1}\matsqrt & \processnoisecov_l\matsqrt
    \end{pmatrix}\transposed,
\end{equation}
and, hence, $(\transitionmat_l\filtcov_{l-1}\matsqrt \q \processnoisecov_l\matsqrt)$ is a matrix square-root of $\predcov_l$.
Similarly, the correction step can be carried out entirely on square-root factors. For more details, see \citet{Grewal2011}.

\subsection{Dynamical low-rank approximation}\label{subsec:background-dlr}
Consider an initial value problem with a matrix-valued flow field $F: \Re \times \Re^{u \times v} \rightarrow \Re^{u \times v}$
\begin{equation}\label{eq:large-mat-diffeq}
    \dot{\mat{M}}(t) = F(t, \mat{M}(t)), \qq \mat{M}(t_0) = \mat{M}_0,
\end{equation}
where $u$ and $v$ are potentially very large.
Dynamic low-rank approximation (DLRA) methods \citep{Koch2007,Ceruti2022} efficiently compute low-rank factorizations $\mat{Y}(t) = \mat{U}(t)\mat{D}(t)\mat{V}\transposed(t) \approx \mat{M}(t)$ of \cref{eq:large-mat-diffeq} by solving
\begin{equation}\label{eq:proj-general-matdiffeq}
    \dot{\mat{Y}}(t) =  \mathcal{P}_r[\mat{Y}(t)] \circ F(t, \mat{Y}(t)), \qq \mat{Y}(t_0) = \mat{U}_0\mat{D}_0\mat{V}_0\transposed,
\end{equation}
instead.
The matrices $\mat{U}_0 \in \Re^{u \times \lowrankdim}$, $\mat{D}_0 \in \Re^{\lowrankdim\times\lowrankdim}$, and $\mat{V}_0 \in \Re^{v \times \lowrankdim}$ are an initial low-rank factorization of $\mat{Y}(t_0) \approx \mat{M}(t_0)$.
$\mathcal{P}_r[\mat{Y}(t)]$ denotes the projection operator onto the tangent space at $\mat{Y}(t)$, where $\mat{Y}(t)$ lies in the manifold of rank-$\lowrankdim$ matrices.
In the present work, we leverage this technique to obtain a low-rank factor $\processnoisecov\matsqrt$ of the process-noise covariance at the next prediction location.
Thus, we avoid much of the computational cost in the prediction step, otherwise caused by solving the full Lyapunov equation for $\processnoisecov$ in \cref{eq:process-noise-cov-diffeq}, for example using matrix-fraction decomposition \citep{Stengel1994,Axelsson2015}.
Concretely, let $F(\processnoisecov) = \driftmat \processnoisecov + \processnoisecov \driftmat\transposed + \dispmat \dispmat\transposed$.
Then, an approximate low-rank process-noise covariance matrix associated with the prediction step is obtained by integrating
\begin{equation}\label{eq:proj-lyapunov-q}
    \dot{\processnoisecov}(t) = \mathcal{P}_r[\processnoisecov(t)] \circ F(\processnoisecov(t)), \qq \processnoisecov(t_{l-1}) = \mat{U}_0\mat{D}^2_0\mat{U}_0\transposed = \mat{0}
\end{equation}
from $t_{l-1}$ to $t_l$. At the start of the filtering recursion $t_0$, an initial low-rank factorization is constructed with a random orthogonal matrix $\lowrank{\mat{U}}_0$ and $\lowrank{\mat{D}}_0 = 0$. 
For all subsequent steps the propagated orthogonal basis $\mat{U}(t_l)$ can be reused.
In this work, all mentions of dynamic low-rank integration always refer to the recently developed, numerically stable basis update \& Galerkin (BUG) integrator \citep{Ceruti2022}, whose error bounds are independent of small singular values. More details are given in \cref{supp:sec-dlra}.

%% file: parts/section_3.tex
\section{Rank-reduced Kalman filtering}\label{sec:method}
In this section, a method for approximate inference in \cref{eq:pomp} is developed.
The idea is to approximate the full filtering/smoothing covariance matrices of the latent state by an
eigendecomposition truncated at the $\lowrankdim$-th largest eigenvalue.
The main challenge is to thereby attain linear computational scaling in $\statedim$ and $\measdim$
under appropriate assumptions on the state-space model. 
Before proceeding with the filtering and smoothing recursions, it is instructive to examine inference in a static low-rank model.

\subsection{Efficient inference in static models of low rank}
Consider the following latent variable model
\begin{equation}\label{eq:static_model}
\vec{x} \sim \N(\vec{\mu}, \predcov\matsqrt  \predcov\transposedsqrt), \qq%
\vec{y} \sim \N(\measurementmat \vec{x}, \measurementnoisecov),
\end{equation}
where $\predcov\matsqrt  \in \Re^{\statedim \times \lowrankdim}$ is of full column rank, $\lowrankdim \leq \statedim$, and $\measurementnoisecov \in \Re^{\measdim \times \measdim}$.
Representing the latent state $\vec{x}$ as
\begin{equation}\label{eq:low_rank_state_model}
\vec{x} = \vec{\mu} + \predcov\matsqrt  \vec{z}, \qq \vec{z} \sim  \N(0, \mathrm{I}_{\lowrankdim\times \lowrankdim}),
\end{equation}
reduces the problem to inference in the following model
\begin{equation}\label{eq:reduced_static_model}
\vec{z} \sim \N(0, \mathrm{I}_{\lowrankdim\times \lowrankdim}),\qq%
\vec{y} \mid \vec{z} \sim \N(\measurementmat \vec{\mu} + \measurementmat\predcov\matsqrt  \vec{z}, \measurementnoisecov).
\end{equation}
An efficient inference scheme is given by the following proposition, which is proved in \cref{supp:subsec-proof-static-model-inference}.
\begin{proposition}\label{prop:static_model_inference}
%
%
Let $\vec{z}$ and $\vec{y}$ be two random variables governed by \cref{eq:reduced_static_model}. Assume $\lowrankdim \leq \measdim$
and consider the following singular value decomposition $(\measurementnoisecov\matsqrtinv \measurementmat \predcov\matsqrt )\transposed = \mat{U} \mat{D} \mat{V}\transposed$,
where $\mat{U}, \mat{D} \in \Re^{\lowrankdim \times \lowrankdim}$ and  $\mat{V} \in \Re^{\measdim \times \lowrankdim}$.
Then, defining the whitened residual $\vec{e} = \measurementnoisecov\matsqrtinv (\vec{y} - \measurementmat\vec{\mu})$, we get
\begin{subequations}
\begin{align}
    \vec{y} &\sim \N(\measurementmat\vec{\mu}, \measurementnoisecov\matsqrt ( \mat{V} \mat{D}^2 \mat{V}\transposed + \mathrm{I} )\measurementnoisecov\transposedsqrt), \\
    \vec{z} \mid \vec{y} &\sim \N( \mat{U}(\mathrm{I} + \mat{D}^2)\matinv \mat{D} \mat{V}\transposed \vec{e}, \mat{U}(\mathrm{I} + \mat{D}^2)\matinv \mat{U}\transposed  ).
\end{align}
\end{subequations}
Furthermore, let\, $\abs{\, \cdot \, }$ denote the matrix determinant. The marginal log-likelihood of $\vec{y}$ is given by
\begin{equation}
\begin{split}
\log \N(\vec{y}; \measurementmat\vec{\mu}, \measurementnoisecov\matsqrt ( \mat{V} \mat{D}^2 \mat{V}\transposed + \mathrm{I} )\measurementnoisecov\transposedsqrt ) &= - \frac{\measdim}{2} \log 2\pi  - \log \abs[0]{\measurementnoisecov\matsqrt } - \frac{1}{2} \sum_{k=1}^\lowrankdim \log(1 + \mat{D}_{kk}^2) \\
&\quad - \frac{1}{2}\norm{\vec{e}}^2 + \frac{1}{2} \vec{e}\transposed \mat{V} \mat{D} (\mat{D}^2  + \mathrm{I})\matinv \mat{D} \mat{V}\transposed \vec{e}.
\end{split}
\end{equation}
\end{proposition}
The below corollary follows from the deterministic relationship between $\vec{z}$ and $\vec{x}$ given by \cref{eq:low_rank_state_model}.

\begin{corollary}\label{corr:static_model_inference}
Let $\vec{y}$ and $\vec{x}$ be two random variables governed by the model \cref{eq:static_model} and $\lowrankdim \leq \measdim$.
Then
\begin{equation}
\vec{x} \mid \vec{y} \sim \N( \vec{\mu} + \mat{K}\vec{e}, \filtcov),
\end{equation}
where
%
\begin{equation}
\mat{K} = \predcov\matsqrt \mat{U}(\mathrm{I} + \mat{D}^2)\matinv \mat{D} \mat{V}\transposed, \qq%
\filtcov = \predcov\matsqrt  \mat{U}(\mathrm{I} + \mat{D}^2)\matinv \mat{U}\transposed \predcov\transposedsqrt.
\end{equation}
%
Moreover, a square-root of $\filtcov$ is readily obtained by
\begin{equation}
    \filtcov\matsqrt  = \predcov\matsqrt  \mat{U}(\mathrm{I} + \mat{D}^2)\matsqrtinv.
\end{equation}
\end{corollary}

\subsection{The rank-reduced filtering recursion}\label{subsec:filtering}
In this section, a low-rank prediction-correction recursion is developed for the purpose of
obtaining the filtering densities and the logarithm of the marginal likelihood.
\paragraph{The prediction equations}
Suppose the filtering distribution at time $t_{l-1}$ is given by
\begin{equation}
p(\vec{x}_{l-1} \mid \vec{y}_{1:l-1}) = \N(\vec{x}_{l-1}; \vec{\mu}_{l-1}, \lowrank{\filtcov}\matsqrt _{l-1} \lowrank{\filtcov}\transposedsqrt_{l-1}),
\end{equation}
where $\lowrank{\filtcov}\matsqrt _{l-1} \in \Re^{\statedim \times \lowrankdim}$.
We begin by detailing how to compute a low-rank factor $\predcov_l\matsqrt$ of the predicted covariance. First compute the truncated singular value decomposition (SVD)
\begin{equation}\label{eq:trunc-svd-predcov}
    \begin{pmatrix}\transitionmat_l\lowrank{\filtcov}_{l-1}\matsqrt & \lowrank{\processnoisecov}\matsqrt_l\end{pmatrix} \approx \tilde{\mat{U}}_l\tilde{\mat{D}}_l\tilde{\mat{V}}_l\transposed
\end{equation}
of the square-root factor (cf.\,\cref{subsec:background-sqrt-kf}),
with $\tilde{\mat{U}}_l \in \Re^{\statedim \times \lowrankdim}, \tilde{\mat{D}}_l \in \Re^{\lowrankdim \times \lowrankdim}$, and $\tilde{\mat{V}}_l \in \Re^{\lowrankdim \times \lowrankdim}$.
This can be done in an optimal way by computing the full SVD and truncating it at the $\lowrankdim$-th largest singular value in $\mathcal{O}(\statedim\lowrankdim^2)$ \citep{golub2013matrix}.
The process-noise covariance factor $\processnoisecov\matsqrt_l \in \Re^{\statedim\times\lowrankdim}$ in \cref{eq:trunc-svd-predcov} is computed using DLRA as described in \cref{subsec:background-dlr}.
The rank-reduced predicted moments at time $t_l$ are thus given as
\begin{align}
    \vec{\mu}^-_l &= \transitionmat_l\mu_{l-1}\q \in \Re^\statedim ,\\
    \lowrank{\predcov}_l\matsqrt &= \tilde{\mat{U}}_l\tilde{\mat{D}}_l \q \in \Re^{\statedim \times \lowrankdim}.
\end{align}
\paragraph{The update equations}
The low-rank update equations follow from \cref{prop:static_model_inference} and \cref{corr:static_model_inference}:
\begin{align}
    \vec{\mu}_l &= \vec{\mu}_l^- + \mat{K}_l \vec{e}_l \q \in \Re^{\statedim}, \\
    \lowrank{\filtcov}_l\matsqrt  &= \lowrank{\predcov}_l\matsqrt  \mat{U}_l ( \mathrm{I} + \mat{D}_l^2)\matsqrtinv \q \in \Re^{\statedim \times \lowrankdim},
\end{align}
with the whitened residual $\vec{e}_l = \measurementnoisecov\matsqrtinv(\vec{y}_l - \measurementmat \vec{\mu}_l^-)$ and
\begin{equation}\label{eq:correct-step-svd}
(\measurementnoisecov\matsqrtinv \measurementmat \lowrank{\predcov}_l\matsqrt )\transposed = \mat{U}_l \mat{D}_l \mat{V}_l\transposed, \qq \mat{K}_l = \lowrank{\predcov}_l\matsqrt  \mat{U}_l (\mathrm{I} + \mat{D}_l^2)\matinv \mat{D}_l \mat{V}_l\transposed,
\end{equation}
for $\mat{U}_l,\mat{D}_l \in \Re^{\lowrankdim\times\lowrankdim}$ and $\mat{V}_l \in \Re^{\measdim\times\lowrankdim}$. The marginal predictive log-likelihood of $\vec{y}_l$ is given by
\begin{equation}\label{eq:marginal-pred-loglik}
\begin{split}
\log p(\vec{y}_l \mid \vec{y}_{1:l-1}) &= - \frac{\measdim}{2} \log 2\pi - \log \abs[0]{\measurementnoisecov\matsqrt }
- \frac{1}{2} \sum_{k=1}^\lowrankdim \log ( (\mat{D}_l)_{kk}^2 + 1)  \\
&\quad - \frac{\norm{\vec{e}_l}^2}{2} + \frac{1}{2} \vec{e}_l\transposed \mat{V}_l \mat{D}_l (\mat{D}_l^2  + \mathrm{I})\matinv \mat{D}_l \mat{V}_l\transposed \vec{e}_l.
\end{split}
\end{equation}
\cref{corr:static_model_inference}---and by extension, the above filtering recursion---is only valid when $\lowrankdim \leq \measdim$, which is exactly the intended usecase for the proposed method.
For settings in which $\measdim < \lowrankdim \leq \statedim$ the correction step follows the square-root correction of a Kalman filter and is detailed in \cref{supp:sec-r-larger-m}.

\subsection{Time complexity of the rank-reduced filtering recursion}\label{subsec:method-complexity}
This section analyzes the computational complexity of the proposed method.
In the worst case, the method scales quadratically in the state dimension $\statedim$ and measurement dimension $\measdim$.
Under favorable conditions, which are met in many real-world applications, the rank-reduced Kalman filter obtains a computational complexity of $O(\statedim \lowrankdim^2)$ as stated by \cref{prop:filter-complexity}, which is proven in \cref{supp:subsec-proof-filter-complexity}.
We begin by specifying a set of assumptions for \cref{prop:filter-complexity}.
\begin{assumption}\label{assumption:cheap-prediction}
    The maps $\vec{x} \mapsto \driftmat \vec{x}$, $\vec{x} \mapsto \transitionmat \vec{x}$, and $\vec{x} \mapsto \dispmat\dispmat\transposed \vec{x}$ can be evaluated in $\mathcal{O}(\statedim)$.
\end{assumption}
\Cref{assumption:cheap-prediction} is fulfilled naturally in many dynamical systems, in which local operators introduce sparsity into the dynamics, as, e.g., in finite-difference approximations of spatial differential operators.
Another example is Kronecker structure in the system matrices arising in spatio-temporal GP regression by assuming the covariance function is separated into a product over the respective spatial and temporal components \citep{Sarkka2012,Solin2016,Hamelijnck2021,Tebbutt2021}.

\begin{assumption}\label{assumption:cheap-observation}
    The map $\vec{x} \mapsto \measurementmat \vec{x}$ can be evaluated in $\mathcal{O}(\measdim)$.
 \end{assumption}

 \begin{assumption}\label{assumption:diagonal-R}
    The map $\vec{x} \mapsto \measurementnoisecov\matsqrtinv \vec{x}$ and the log-determinant $\log\abs[0]{\measurementnoisecov\matsqrt}$ can be evaluated in $\mathcal{O}(\measdim)$.
 \end{assumption}

 Throughout this work, we refer to the situation in which \cref{assumption:cheap-prediction} or \ref{assumption:cheap-observation} do not apply as the ``worst case''.
 \Cref{assumption:diagonal-R} is taken for granted as the measurement-noise covariance $\measurementnoisecov$ is often a diagonal matrix, which implies that sensor errors are uncorrelated. This is not only realistic but also commonly imposed in modelling.
The above assumptions allow the following proposition.
\begin{proposition}\label{prop:filter-complexity}
    Given \crefrange{assumption:cheap-prediction}{assumption:diagonal-R}, the proposed method approximates the filtering densities and the marginal likelihood at a cost of
    $\mathcal{O}(\statedim \lowrankdim^2 + \measdim\lowrankdim^2 + \lowrankdim^3)$.
\end{proposition}
The below corollary follows from the book-keeping in the proof of \cref{prop:filter-complexity} in \cref{supp:subsec-proof-filter-complexity}. 
\begin{corollary}\label{corr:filter-worst-case-complexity}
    When \cref{assumption:cheap-prediction} or \cref{assumption:cheap-observation} are not satisfied, the worst-case complexity of the proposed low-rank filtering recursion is $\mathcal{O}(\statedim^2 \lowrankdim + \statedim\measdim + \measdim^2\lowrankdim)$.
    %
\end{corollary}
Overall, the proposed filtering recursions achieve the same asymptotic complexity as the existing ensemble methods under the same set of assumptions.
However, our method is entirely deterministic.
%

\subsection{The rank-reduced smoothing recursion}\label{subsec:smoothing}
It remains to obtain a recursion for the smoothing densities,
which can be shown to be computationally tractable given that the filtering recursion is tractable.
The smoothing densities are denoted by
\begin{equation}
p(\vec{x}_l \mid \vec{y}_{1:\numT}) = \N(\vec{x}_l; \xi_l, \Lambda_l), \quad \xi_\numT = \vec{\mu}_\numT, \ \Lambda_\numT = \filtcov_\numT.
\end{equation}
It can be shown that the posterior process has a backward Markov representation \citep{Cappe2005}, hence the smoothing marginals may be obtained by
\begin{equation}\label{eq:backward_recursion}
    p(\vec{x}_l \mid \vec{y}_{1:\numT}) = \int b_{l,l+1}(\vec{x}_l \mid \vec{x}_{l+1}) p(\vec{x}_{l+1} \mid \vec{y}_{1:\numT}) \dif \vec{x}_{l+1},
\end{equation}
where we call $b_{l,l+1}$ the \emph{backwards kernel}. Consequently, the problem consists of approximating $b_{l,l+1}$ such that the marginalization \cref{eq:backward_recursion} may
be implemented in a computationally frugal manner.
\paragraph{Approximating the backwards kernel}
For the linear Gaussian case, the backward kernel \citep{Tronarp22}
\begin{equation}
    b_{l, l+1}(\vec{x}_{l} \mid \vec{x}_{l+1}) = \N(\vec{x}_{l}; \mat{G}_l\vec{x}_{l+1} + \vec{v}_l, \mat{P}_l),
\end{equation}
is parametrized by the \emph{smoothing gain} $\mat{G}_l \in \Re^{\statedim\times\statedim}$, the shift vector $\vec{v}_l \in \Re^{\statedim}$, and the covariance of the backwards kernel $\mat{P}_l \in \Re^{\statedim\times\statedim}$. Let $(\cdot)\matpseudoinv$ denote the Moore--Penrose pseudoinverse. Then
\begin{equation}
    \mat{G}_l = \filtcov_{l}\transitionmat_{l+1}\transposed\predcov_{l+1}\matpseudoinv, \quad\vec{v}_l = \mu_{l} - \mat{G}_l\mu^-_{l+1},\quad
    \mat{P}_l = (\mat{I} - \mat{G}_l\transitionmat_{l+1}) \filtcov_{l} (\mat{I} - \mat{G}_l\transitionmat_{l+1})\transposed + \mat{G}_l\processnoisecov_{l+1}\mat{G}_l\transposed.
\end{equation}
In the low-rank setting, the backwards kernel is efficiently approximated given the results from above.
The approximate smoothing gain is a product of a tall, a small quadratic, and a wide matrix:
\begin{equation}\label{eq:approx-smoothing-gain}
    \mat{G}_l \approx \lowrank{\filtcov}_{l}\matsqrt \underbrace{\lowrank{\filtcov}_{l}\transposedsqrt\transitionmat_{l+1}\transposed\left(\lowrank{\predcov}_{l+1}\matsqrt\right)\matpseudoinv}_{=: \mat{\Gamma}_l \in \Re^{\lowrankdim\times\lowrankdim}}\left(\lowrank{\predcov}_{l+1}\transposedsqrt\right)\matpseudoinv.
\end{equation}
During filtering, $\mat{\Gamma}_l$ is saved alongside the low-rank factors of the prediction and filtering covariances.
We proceed to compute a low-rank representation of the backwards-transition covariance.
Consider the following singular value decomposition, truncated at the $\lowrankdim$-th largest singular value,
\begin{equation}\label{eq:backwards-kernel-process-noise-cov-factorization}
    \begin{pmatrix}
        (\mat{I} - \mat{G}_l\transitionmat_{l+1}) \lowrank{\filtcov}_{l}\matsqrt & \mat{G}_l\lowrank{\processnoisecov}\matsqrt_{l+1}
    \end{pmatrix}
    \approx \widehat{\mat{U}}_l\widehat{\mat{D}}_l\widehat{\mat{V}}_l\transposed,
\end{equation}
where $\widehat{\mat{U}}_l \in \Re^{\statedim \times \lowrankdim}$, $\widehat{\mat{D}}_l \in \Re^{\lowrankdim \times \lowrankdim}$ diagonal, $\widehat{\mat{V}}_l \in \Re^{\lowrankdim \times \lowrankdim}$, and $\mat{G}_l$ as given in \cref{eq:approx-smoothing-gain}.
Then a rank-$\lowrankdim$ factor of $\mat{P}_l$ is given as
\begin{equation}\label{eq:backwards-kernel-process-noise-cov}
    \lowrank{\mat{P}}_l\matsqrt = \widehat{\mat{U}}_l\widehat{\mat{D}}_l \in \Re^{\statedim \times \lowrankdim}.
\end{equation}

As for filtering, in the worst case, the cost for low-rank smoothing scales quadratically with the state dimension $\statedim$.
The following proposition, which is proved in \cref{supp:subsec-proof-smoothing-complexity}, states that this can be further reduced to linear complexity in $\statedim$, given the assumptions on the dynamics model
are satisfied.
\begin{proposition}\label{prop:smoothing-complexity}
    Given \cref{assumption:cheap-prediction}, computing
    the smoothing density $p(\vec{x}_l \mid \vec{y}_{1:\numT})$ costs $\mathcal{O}(\statedim\lowrankdim^2 + \lowrankdim^3)$.
\end{proposition}
Finally, it is useful to mention that realizations of the backwards process are posterior samples.

%% file: parts/section_4.tex
\section{Experiments}\label{sec:experiments}


This section evaluates the proposed method in different experimental settings.
The chosen measure of quality is the distance of the approximate low-rank moments to the exact KF.
We measure the mean deviations with the root-mean-squared error (RMSE) and the covariance deviations with the time-averaged relative Frobenius distances.
The presented rank-reduced Kalman filter (RRKF) is compared to the ensemble Kalman filter (EnKF) and the ensemble transform Kalman filter (ETKF).
All EnKF and ETKF results are given in sample statistics over 20 runs.
\Cref{subsec:LA-exp} shows that for a truly low-rank system, our method recovers
the KF estimate up to numerical error as soon as $\lowrankdim$ exceeds the true rank of the problem.
\Cref{subsec:london-exp} tests the method on a spatio-temporal GP regression problem with real data.
After \cref{subsec:on-model-exp} evaluates the approximation quality for increasingly low-rank systems in a controlled experimental environment,
\cref{subsec:runtime-exp} verifies the stated asymptotic cost of the method.
Finally, a large-scale spatio-temporal regression problem is solved in \cref{subsec:australia-exp}. 
In all experiments, we compute the stationary mean and a low-rank factorization of the stationary covariance matrix of the prior and condition the stationary moments on the first measurement in the respective time-series dataset.
More details on the experimental setups are given in \cref{supp:sec-exp-details}.

\begin{figure}
  \centering
  \begin{subfigure}[b]{0.42\linewidth}
      \centering
      \includegraphics[width=\linewidth]{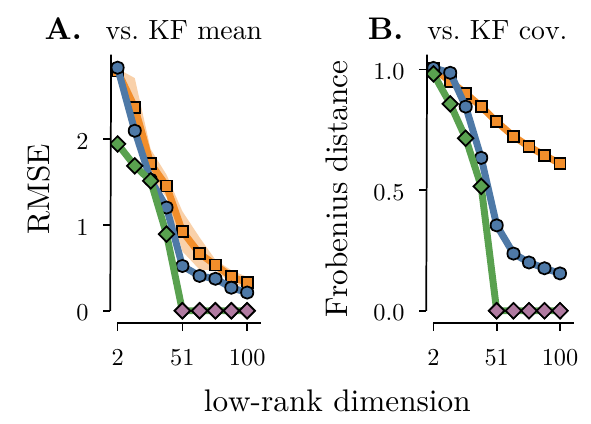}
      \caption{\emph{Linear advection dynamics (true rank = 51).}}
      \label{fig:LA-subplot}
  \end{subfigure}
  \hfill
  \begin{subfigure}[b]{0.52\linewidth}
      \centering
      \includegraphics[width=\linewidth]{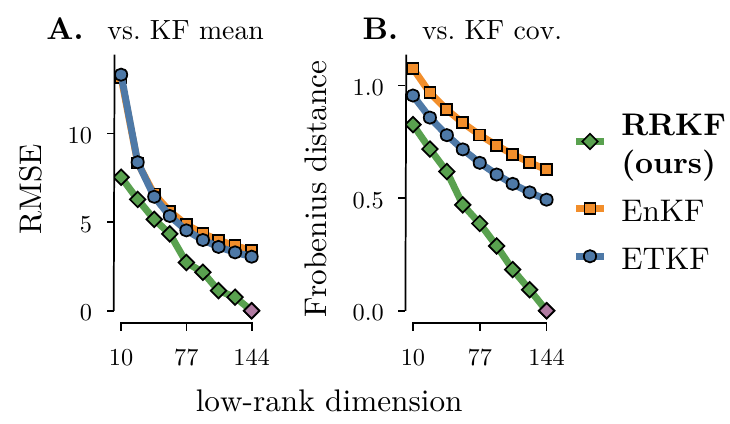}
      \caption{\emph{London air-quality regression.}}
      \label{fig:london-subplot}
  \end{subfigure}
  \caption{
    \emph{Performance of the low-rank filters on two different settings.}
    In the truly low-rank linear advection problem (a), the RRKF achieves the optimal estimate (\kfmarker[0.5ex]{purple}) when $\lowrankdim$ exceeds the true rank of the problem.
    For the spatio-temporal air-quality regression (b), our method is consistently better. 
    }
  \label{fig:LA-and-london-exp}
\end{figure}

\subsection{Linear advection model}\label{subsec:LA-exp}
The proposed algorithm is evaluated in a standard data-assimilation setup in which linear-advection
dynamics with periodic boundary conditions are assimilated to a set of simulated data.
The setup follows the description in \citet{Sakov2008} and the experimental details are additionally detailed in \cref{supp:sec-exp-details}.
This problem has rank 51 by construction.
\Cref{fig:LA-subplot} shows how the exact KF estimate is recovered by our method for $\lowrankdim = 51$, while the ensemble methods converge according to a Monte--Carlo rate.
This is particularly clear for the covariance estimate (\cref{fig:LA-subplot}, \textbf{B.}).

\subsection{London air-quality regression}\label{subsec:london-exp}

This experiment uses hourly data from the London air-quality network \citep{data_laqn} between January 2019 and April 2019, which amounts to measurements at $72$ spatial locations and 2159 points in time.
The data used, together with its processing (except the train-test split), is the same as in the corresponding experiment by \citet{Hamelijnck2021}.
The model is a spatio-temporal Mat\'ern-\nicefrac{3}{2} process with prior hyperparameters that maximize the marginal log likelihood.
\Cref{fig:london-subplot} shows that---compared to the ensemble methods---our algorithm is consistently closer to the KF estimate and recovers it at $\lowrankdim = \statedim$.


\begin{figure}
  \centering
  \includegraphics[width=.99\linewidth]{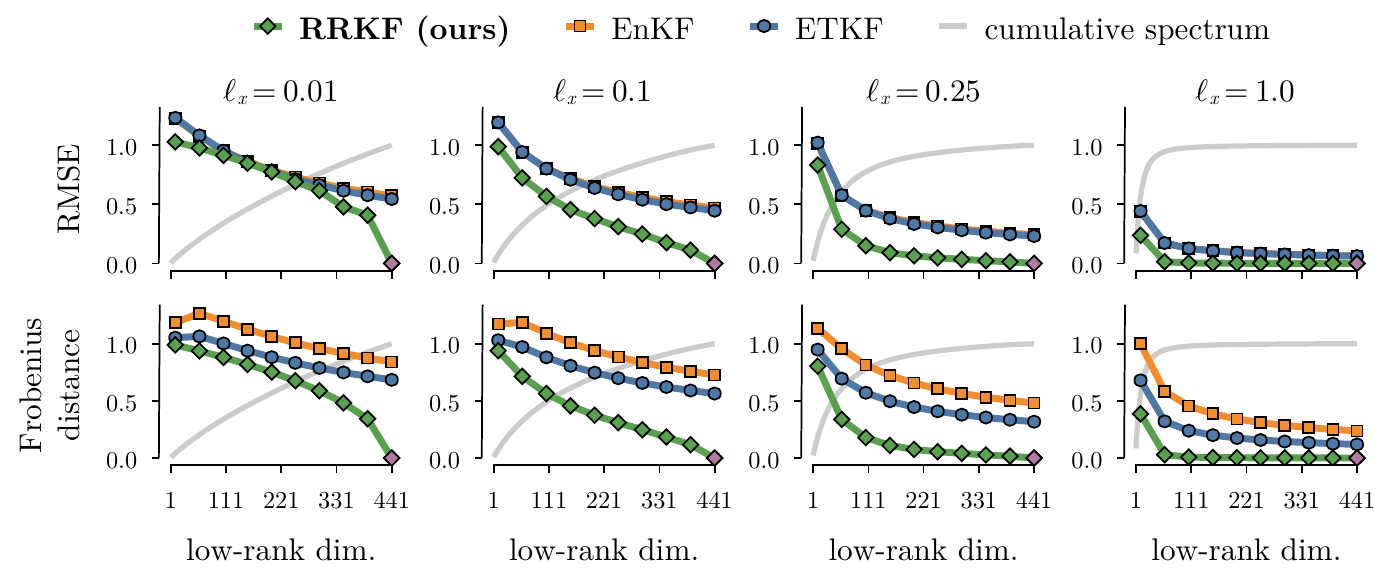}
  \caption{
  \emph{How does the rate of spectral decay influence the low-rank approximation?}
  The larger the spatial length scale of a spatio-temporal Mat\'ern model, the faster the spectrum decays and the faster all methods converge to the KF estimate.
  The first and second row compare the mean estimates and the covariance estimates, respectively.
  The RRKF estimate is consistently closer and recovers the KF estimate (\kfmarker[0.5ex]{purple}) at $\lowrankdim = \statedim$.
  The cumulative spectrum of the final-step KF covariance is shown in grey.
  }
  \label{fig:different_lx_r_vs_error}
\end{figure}
\subsection{Spatio-temporal Mat\'ern process with varying spatial lengthscale}\label{subsec:on-model-exp}
Consider a spatio-temporal Gaussian process $\vec{x}(t) \sim \mathcal{GP}(0, k_t \otimes k_x)$ with covariance structure that is separable in time and space.
Let both $k_t$ and $k_x$ be of the Mat\'ern family with characteristic spatial and temporal lengthscales $\ell_t, \ell_x$ and output scales $\sigma^2_t, \sigma^2_x$.
Spatio-temporal Mat\'ern processes can be translated to the formulation in \cref{eq:pomp_dynamics}, as detailed, for instance, by \citet{Solin2016}.
This setting allows for varying the ``low-rankness'' by the choice of $\ell_x$: the larger $\ell_x$, the more information about a spatial point is
prescribed by its surroundings.
As $\ell_x$ approaches zero, the spatial kernel matrix gets more and more diagonal, encoding spatially independent states,
and causing a barely decaying spectrum.
The spectrum for large $\ell_x$ decays rapidly, making low-rank approximations more accurate for small $\lowrankdim$.

In the experiment, we consider a spatial domain $[0, 2] \times [0, 2] \subset \Re^2$, which is subsampled at a uniformly-spaced grid with $\Delta_x = 0.1$ ($\statedim = 21 \times 21 = 441$).
Noisy observations of the full state trajectory, are drawn from a realization of the prior process.
\Cref{fig:different_lx_r_vs_error} demonstrates that, for increasing spatial lengthscales, all methods converge faster to the true KF estimate.
Crucially, however, our method is consistently closer to the optimal estimate and recovers it for $\lowrankdim = \statedim$.

\begin{figure}
  \centering
  \includegraphics[width=.99\linewidth]{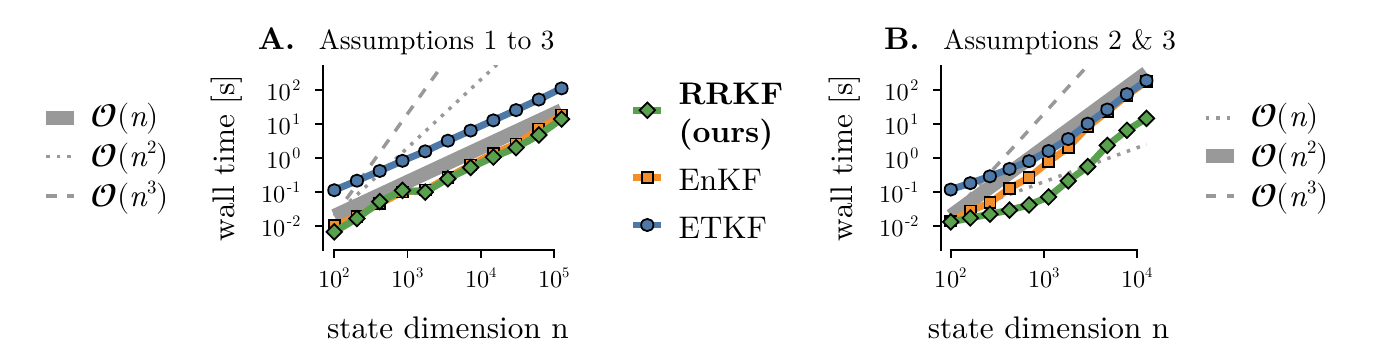}
  \caption{
    \emph{Best-case and worst-case asymptotic complexities of the low-rank filters.}
    When all \crefrange{assumption:cheap-prediction}{assumption:diagonal-R} are satisfied (\textbf{A.}), the elapsed time grows linearly with the state dimension $\statedim$.
    If parts of the assumptions are not met (\textbf{B.}), the cost scales quadratically with the respective dimension.
  }
  \label{fig:time_plot}
\end{figure}

\subsection{Runtime}\label{subsec:runtime-exp}

To demonstrate the asymptotic computational complexities given in \cref{subsec:method-complexity}, we
investigate the runtimes on two different settings: The first fulfills both \cref{assumption:cheap-prediction,assumption:cheap-observation}, whereas
the second only fulfills \cref{assumption:cheap-observation}, which means that the cost of prediction is quadratic in the state dimension.
Both problems are solved on a temporal grid of size $\numT=100$ with the measurement dimension fixed at $\measdim = 100$ and the low-rank dimension fixed at $r = 5$.

The problem from \cref{subsec:LA-exp} serves as the best-case setting.
The noise-free linear-advection dynamics amount to multiplication of the state with a circulant matrix, which
is implemented efficiently using fast Fourier transform. The measurement operator amounts to array indexing.
As subplot \textbf{A.} in \cref{fig:time_plot} confirms, the computational cost grows linearly with the state dimension.
As a worst-case example we employ a spatio-temporal Mat\'ern process with a dense spatial kernel matrix (cf.\,\cref{subsec:on-model-exp}).
In this case, we expect the computation time to scale quadratically with the state dimension because $\dispmat\dispmat\transposed$ is dense. This is verified by the right plot in \cref{fig:time_plot} (\textbf{B.}).

\subsection{Large-scale spatio-temporal GP regression on rainfall data}\label{subsec:australia-exp}

\begin{figure}
  \centering
  \includegraphics{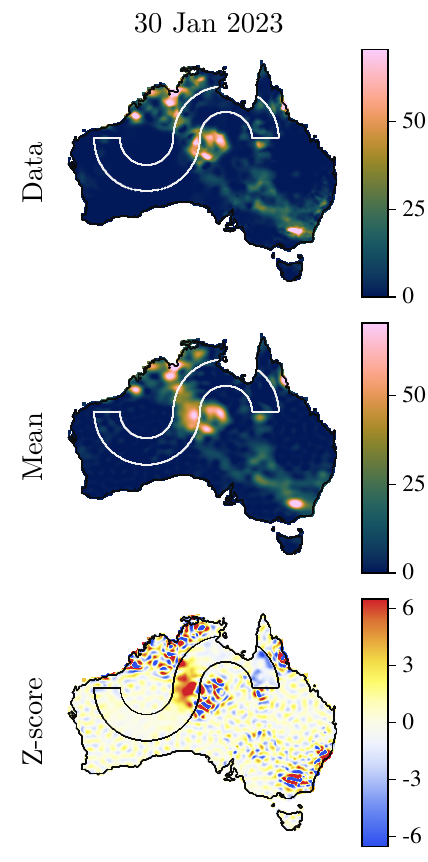}\hfill%
  \includegraphics{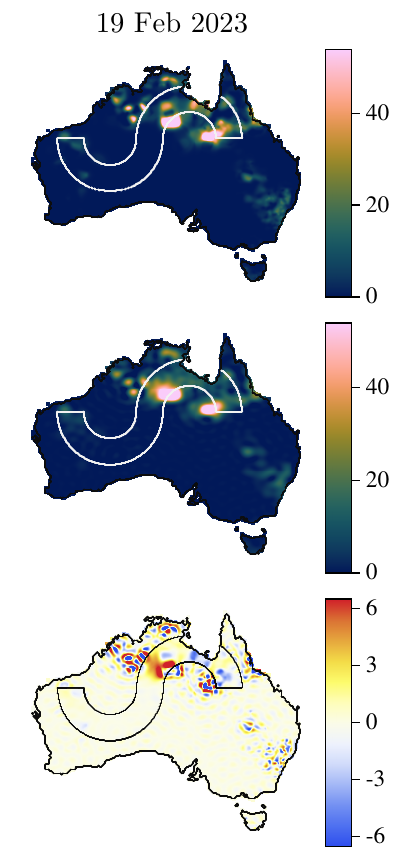}\hfill%
  \includegraphics{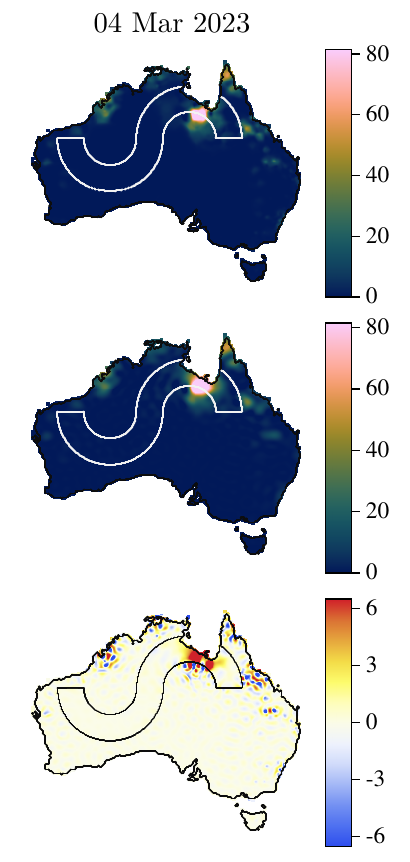}%
  \caption{
    \textit{Rainfall in Australia.}
    A spatio-temporal GP regression problem with $\statedim = \australiastatedim$ state dimensions, solved
    with the RRKF with $\lowrankdim = 1000$.
    The columns are three different days in the time series.
    The rows show the data, mean estimate, and Z-scores, respectively.
  }
  \label{fig:australia_rainfall}
\end{figure}

As a final experiment, a spatio-temporal GP regression problem is solved
on a large gridded analysis data set measuring rainfall in Australia. The data is provided by the Australian Water Availability Project (AWAP) \citep{AWAPdata,Jones2009}.
The number of spatial data points is $\statedim = \australiastatedim$, which rules out the cubic-in-$\statedim$ Kalman filter.
The smoothing posterior is computed at $\numT = 40$ time points, from 25 January, 2023 through 5 March, 2023.
The total number of observations is thus $\statedim \cdot \numT = 1\,236\,440$.
The spatio-temporal prior Mat\'ern model is selected by maximizing the marginal log-likelihood of a lower-resolution data set.
A set of $6160$ spatial points is excluded from the data during filtering and smoothing in order to evaluate spatial interpolation performance.
Thus, the measurement dimension is $\measdim = \statedim - 6160 = 24751$.
\Cref{fig:australia_rainfall} shows the low-rank approximation to the GP posterior with $\lowrankdim = 1000$ low-rank dimensions.
The non-observed spatial locations are framed in the horizontal-``S'' shape.
At measurement locations, the data is described well by the model, while a slight smoothing-out effect can be observed due to the high-frequency features of the model being truncated.
At the evaluation points the interpolation aligns with the expectations regarding the simple Mat\'ern model.
The Z-score map shows $\frac{\xi_l - \vec{y}_l}{\lambda_l}$, where $\xi_l$ and $\lambda_l$ denote the smoothing mean and marginal standard deviation and $\vec{y}_l$ the data point, at time $t_l$.
This highlights badly calibrated uncertainty estimates, wherever the model under-/overestimates the data and is divided by a small standard deviation, in blue/red.
The Z-score distribution receives further attention in \cref{supp:sec-calibration}.
All in all, the RRKF achieves high-quality approximate estimates while compressing the state by a factor of $\frac{\australiastatedim}{1000} \approx 30$.

%% file: parts/section_5.tex
\section{Limitations}\label{sec:limitations}
The main limitation of the proposed approximate probabilistic inference scheme is that the truncation to $\lowrankdim$ dimensions cuts away covariance information, which is not accounted for.
As a result, confidence \emph{grows} as $\lowrankdim$ shrinks, while estimates should arguably get more uncertain instead with increased compression.
In existing low-rank filters, this issue is often counteracted manually by inflating the covariance matrices, on a per-application basis \citep[Section 4.4]{Carrassi2018} and calls for a more principled treatment.
Finding a way to keep track of that residual uncertainty information is beyond the scope of this paper. It remains unclear how to preserve the stated asymptotic complexities while doing so.

Further, all results herein are stated for linear dynamics and observation models. Extensions to non-linear models could include linearization of the according transitions, or cubature methods \citep{Sarkka2013}.

%% file: parts/conclusion.tex
\section{Conclusion}\label{sec:conclusion}
By building upon well-established knowledge about optimal compression, we have proposed an algorithm
providing a principled way to balance approximation accuracy against computational cost in high-dimensional state estimation.
It combines simple (truncated) singular value decompositions with recent numerical low-rank integrators of large
matrix differential equations.
We have offered both theoretical and empirical arguments for why it is desirable to use a deterministic algorithm when approximating large-scale probabilistic state-estimation problems.

%% file: supplement.tex
\appendixtitle

\appendix
\numberwithin{equation}{section}
\numberwithin{proposition}{section}
\numberwithin{remark}{section}
\numberwithin{figure}{section}
\numberwithin{table}{section}

\section{Proofs}

\subsection{Proof of Proposition \ref{prop:static_model_inference}}\label{supp:subsec-proof-static-model-inference}

\begin{proof}
  By standard results on Gaussian conditioning and marginalization
  \begin{subequations}
  \begin{align}
  \vec{z} \mid \vec{y} &\sim \mathcal{N}( \mat{K}_\vec{z}(\vec{y} - \measurementmat\mu), \filtcov_\vec{z}), \\
  \vec{y} &\sim \mathcal{N}( \measurementmat\mu, \mat{S}),
  \end{align}
  \end{subequations}
  where
  \begin{subequations}
  \begin{align}
  \filtcov_\vec{z}\matinv &= \mathrm{I} + (\measurementnoisecov\matsqrtinv \measurementmat \predcov\matsqrt)\transposed \measurementnoisecov\matsqrtinv C \predcov\matsqrt, \\
  \mat{K}_\vec{z} &= \filtcov_z (\measurementnoisecov\matsqrtinv \measurementmat \predcov\matsqrt)\transposed \measurementnoisecov\matsqrtinv, \\
  \measurementnoisecov\matsqrtinv \mat{S} \measurementnoisecov\matsqrtinvtransposed   &= \measurementnoisecov\matsqrtinv \measurementmat \predcov\matsqrt (\measurementnoisecov\matsqrtinv \measurementmat \predcov\matsqrt)\transposed + \mathrm{I}.
  \end{align}
  \end{subequations}
  Substituting for the singular value decomposition of $(\measurementnoisecov\matsqrtinv \measurementmat \predcov\matsqrt)\transposed$ gives the conditional covariance
  \begin{equation}
  \filtcov_\vec{z} = \big(\mathrm{I} + \mat{U} \mat{D}^2 \mat{U}\transposed\big)\matinv = \big( \mat{U}(\mathrm{I} + \mat{D}^2)\mat{U}\transposed \big)\matinv  = \mat{U}(\mathrm{I} + \mat{D}^2)\matinv\mat{U}\transposed,
  \end{equation}
  the Kalman gain
  \begin{equation}
  \mat{K}_\vec{z} =  \mat{U}(\mathrm{I} + \mat{D}^2)\matinv\mat{U}^* \mat{U} \mat{D} \mat{V}^* \measurementnoisecov\matsqrtinv =  \mat{U}(\mathrm{I} + \mat{D}^2)\matinv \mat{D} \mat{V}^* \measurementnoisecov\matsqrtinv,
  \end{equation}
  and the marginal measurement covariance matrix
  \begin{equation}
  \measurementnoisecov\matsqrtinv \mat{S} \measurementnoisecov\matsqrtinvtransposed = \mat{V} \mat{D}^2 \mat{V}\transposed + \mathrm{I}.
  \end{equation}
  This gives the result on marginalization and conditioning.
  To obtain the expression for the marginal likelihood, which is given by
  \begin{equation}
  \mathcal{N}(\vec{y}; \measurementmat\mu, \mat{S}) = - \frac{\measdim}{2} \log 2\pi - \frac{1}{2} \log \abs{\mat{S}} - \frac{1}{2}(\vec{y} - \measurementmat\mu)\transposed \mat{S}\matinv (\vec{y} - \measurementmat\mu).
  \end{equation}
  The log determinant is given by
  \begin{equation}
  \log \abs{\mat{S}} = \log \abs{\measurementnoisecov\matsqrt(  \mat{V} \mat{D}^2 \mat{V}^* + \mathrm{I}  ) \measurementnoisecov\transposedsqrt} = 2 \log \abs[0]{\measurementnoisecov\matsqrt} + \log \abs[0]{ \mat{V} \mat{D}^2 \mat{V}^* + \mathrm{I}}.
  \end{equation}
  Let $\mat{V}_k$ be the $k$-th column vector of $\mat{V}$ then it is an eigenvector of $ \mat{V} \mat{D}^2 \mat{V}^* + \mathrm{I}$ with eigenvalue $\mat{D}_{kk}^2 + 1$.
  Furthermore, any vector in the orthogonal complement to the column space of $\mat{V}$ is also an eigenvector with eigenvalue $1$. Therefore,
  \begin{equation}
  \log \abs[0]{ \mat{V} \mat{D}^2 \mat{V}^* + \mathrm{I}} = \sum_{k=1}^r \log(\mat{D}_{kk}^2 + 1).
  \end{equation}
  It remains to obtain the desired expression for the quadratic form. Start by inserting the expression for $\mat{S}$
  \begin{equation}
  \begin{split}
  (\vec{y} - \measurementmat\mu)\transposed \mat{S}\matinv (\vec{y} - \measurementmat\mu) &= (\vec{y} - \measurementmat\mu)\transposed \measurementnoisecov\matsqrtinvtransposed (\mat{V} \mat{D}^2 \mat{V}\transposed + \mathrm{I})\matinv \measurementnoisecov\matsqrtinv(\vec{y} - \measurementmat\mu) \\
  &= \vec{e}\transposed  (\mat{V} \mat{D}^2 \mat{V}\transposed + \mathrm{I})\matinv \vec{e}.
  \end{split}
  \end{equation}
  Now $\vec{e}^*  (\mat{V} \mat{D}^2 \mat{V}^* + \mathrm{I})\matinv \vec{e} = \vec{e}^*\vec{b} $, where we define the vector $\vec{b}$ such that
  \begin{equation}
  (\mat{V} \mat{D}^2 \mat{V}^* + \mathrm{I})\vec{b} = \vec{e}.
  \end{equation}
  Multiplying from the left by $\mat{V}^*$ gives
  \begin{equation}
  (\mat{D}^2  + \mathrm{I})\mat{V}^*\vec{b} = \mat{V}^*\vec{e},
  \end{equation}
  therefore,
  \begin{equation}
  \mat{V}^*\vec{b} = (\mat{D}^2  + \mathrm{I})\matinv \mat{V}^*\vec{e}.
  \end{equation}
  Inserting this in the original definition of $\vec{b}$ gives
  \begin{equation}
  \mat{V} \mat{D}^2 (\mat{D}^2  + \mathrm{I})\matinv \mat{V}^*\vec{e} + \vec{b} = \vec{e},
  \end{equation}
  from which it follows that
  \begin{equation}
  \vec{e}^*  (\mat{V} \mat{D}^2 \mat{V}^* + \mathrm{I})\matinv \vec{e} = \vec{e}^*\vec{b}  = \norm{\vec{e}}^2 - \vec{e}^* \mat{V} \mat{D} (\mat{D}^2  + \mathrm{I})\matinv \mat{D} \mat{V}^*\vec{e}.
  \end{equation}
\end{proof}

\subsection{Proof of Proposition \ref{prop:filter-complexity}}\label{supp:subsec-proof-filter-complexity}

In the following, the computational complexity of the low-rank filtering recursion is analyzed in detail, which proves
\cref{prop:filter-complexity} and \cref{corr:filter-worst-case-complexity}.
In the best case, the cost of the proposed filtering algorithms scales linearly in the state dimension $\statedim$ and the measurement dimension $\measdim$.
For this to hold, we assume that
\begin{enumerate}[label=(\alph*)]
  \item the maps $\vec{x} \mapsto \driftmat \vec{x}$, $\vec{x} \mapsto \transitionmat \vec{x}$, and $\vec{x} \mapsto \dispmat\dispmat\transposed \vec{x}$ can be evaluated in $\mathcal{O}(\statedim)$,
  \item the map $\vec{x} \mapsto \measurementmat \vec{x}$ can be evaluated in $\mathcal{O}(\measdim)$, and
  \item the map $\vec{x} \mapsto \measurementnoisecov\matsqrtinv \vec{x}$ and the log-determinant $\log\abs[0]{\measurementnoisecov\matsqrt}$ can be evaluated in $\mathcal{O}(\measdim)$.
\end{enumerate}
%
We refer to the situation in which (a) or (b) do not apply as the ``worst case''.
Assumption (c) is taken for granted as the measurement-noise covariance $\measurementnoisecov$ is often a diagonal matrix, which implies that sensor errors are uncorrelated. This is not only realistic but also commonly imposed in modelling.

\begin{proof} The best-case cost of the approximate filtering scales linear in the state dimension $\statedim$ and the measurement dimension $\measdim$.
In the worst case, the complexity scales quadratically in $\statedim$ and $\measdim$.
\paragraph{Prediction}
We begin by analyzing the cost for the approximate integration of the low-rank process-noise covariance $\processnoisecov\matsqrt$.
This amounts to the cost of the dynamical low-rank approximation (DLRA) algorithm for a symmetric Lyapunov equation.
\Cref{supp:sec-dlra} gives a detailed description of how this algorithm is used as part of the RRKF recursions.
Let the matrix-valued flow field be $F(\processnoisecov) = \driftmat\processnoisecov + \processnoisecov\driftmat\transposed + \dispmat\dispmat\transposed$.
Given an initial factorization $\processnoisecov_0 \approx \mat{Y}_0 =  \mat{U}_0\mat{D}_0^2\mat{U}_0\transposed$, with $\mat{U}_0 \in \Re^{\statedim\times\lowrankdim}$ and $\mat{D}_0 \in \Re^{\lowrankdim\times\lowrankdim}$, the cost for integrating this matrix equation using DLRA is as follows:

\begin{table}
  \caption{Time complexity of reduced-rank filtering: The prediction step}
  \label{table:prediction-complexity}
  \centering
  \begin{tabular}{llll}
    \toprule
    \multicolumn{2}{c}{Step}\\
    \cmidrule(r){1-2}
    Eq. & Operation & Best case & Worst case\\
    \midrule
    \multirow{3}{*}{\eqref{eq:k-step-flowfield-complexity}} & $\driftmat\mat{K}(t)$ & $\mathcal{O}(\statedim\lowrankdim)$  & $\mathcal{O}(\statedim^2\lowrankdim)$  \\
                         & $\mat{K}(t)(\mat{U}_0\transposed(\driftmat\transposed\mat{U}_0))$ & $\mathcal{O}(\statedim\lowrankdim^2)$  & $\mathcal{O}(\statedim^2\lowrankdim + \statedim\lowrankdim^2)$ \\
                         & $\dispmat\dispmat\transposed\mat{U}_0$ & $\mathcal{O}(\statedim\lowrankdim)$  & $\mathcal{O}(\statedim^2\lowrankdim)$ \\
                         \midrule
    \eqref{eq:k-step-qr-complexity} & $\operatorname{QR}\left(\mat{K}(t+h)\right)$ & $\mathcal{O}(\statedim\lowrankdim^2)$  & $\mathcal{O}(\statedim\lowrankdim^2)$  \\
    \midrule
    \eqref{eq:k-step-m-complexity} & $\mat{M} = \mat{U}_h\transposed\mat{U}_0$ & $\mathcal{O}(\statedim\lowrankdim^2)$  & $\mathcal{O}(\statedim\lowrankdim^2)$  \\
    \midrule
    \eqref{eq:s-step-initial-d-complexity} & $\mat{D}(t_0) = \mat{M}\mat{D}_0\mat{M}\transposed$ & $\mathcal{O}(\lowrankdim^3)$  & $\mathcal{O}(\lowrankdim^3)$  \\
    \midrule
    \multirow{3}{*}{\eqref{eq:s-step-flowfield-complexity}} & $\mat{U}_h\transposed(\driftmat(\mat{U}_h\mat{D}(t)))$ & $\mathcal{O}(\statedim\lowrankdim^2)$  & $\mathcal{O}(\statedim^2\lowrankdim + \statedim\lowrankdim^2)$  \\
                         & $(\mat{D}(t)\mat{U}_h\transposed)(\driftmat\transposed\mat{U}_h)$ & $\mathcal{O}(\statedim\lowrankdim^2)$  & $\mathcal{O}(\statedim^2\lowrankdim + \statedim\lowrankdim^2)$ \\
                         & $\mat{U}_h\transposed(\dispmat\dispmat\transposed\mat{U}_h)$ & $\mathcal{O}(\statedim\lowrankdim^2)$  & $\mathcal{O}(\statedim^2\lowrankdim + \statedim\lowrankdim^2)$ \\
    \midrule
    \eqref{eq:post-dlra-sqrt-d} & $\mat{D}_l = \mat{D}_l\matsqrt\mat{D}_l\transposedsqrt$ & $\mathcal{O}(\lowrankdim^3)$ & $\mathcal{O}(\lowrankdim^3)$ \\
    \midrule
    \eqref{eq:predict-mean-complexity} & $\transitionmat_l\mu_{l-1}$ & $\mathcal{O}(\statedim)$ & $\mathcal{O}(\statedim^2)$ \\
    \midrule
    \multirow{2}{*}{\eqref{eq:predict-cov-complexity}} & $\transitionmat_l\filtcov_{l-1}\matsqrt$ & $\mathcal{O}(\statedim\lowrankdim)$ & $\mathcal{O}(\statedim^2\lowrankdim)$ \\
                                             & $(\transitionmat_l\lowrank{\filtcov}_{l-1}\matsqrt \q \lowrank{\processnoisecov}\matsqrt_l) \approx \tilde{\mat{U}}_l\tilde{\mat{D}}_l\tilde{\mat{V}}_l\transposed$ & $\mathcal{O}(\statedim\lowrankdim^2)$ & $\mathcal{O}(\statedim\lowrankdim^2)$ \\
    \bottomrule
  \end{tabular}
\end{table}

\begin{enumerate}
  \item \textbf{K-step}
  \begin{enumerate}
    \item Flow-field evaluation%
    \begin{subequations}\label{eq:k-step-flowfield-complexity}
      \begin{align}
        F(t, \mat{K}(t)\mat{U}_0\transposed)\mat{U}_0
        &= \driftmat(\mat{K}(t)\mat{U}_0\transposed)\mat{U}_0 + \mat{K}(t)\mat{U}_0\transposed\driftmat\transposed\mat{U}_0 + \dispmat\dispmat\transposed\mat{U}_0 \\
        &= \driftmat\mat{K}(t) + \mat{K}(t)(\mat{U}_0\transposed(\driftmat\transposed\mat{U}_0)) + \dispmat\dispmat\transposed\mat{U}_0,
      \end{align}
    \end{subequations}
    \item QR factorization of tall $\statedim\times\lowrankdim$ matrix%
    \begin{equation}\label{eq:k-step-qr-complexity}
      \operatorname{QR}\left(\mat{K}(t+h)\right), 
    \end{equation}
    \item Compute \begin{equation}\label{eq:k-step-m-complexity}
      \mat{M} = \mat{U}_h\transposed\mat{U}_0. 
    \end{equation}
  \end{enumerate}
  \item \textbf{S-step}
  \begin{enumerate}
    \item Compute initial $\mat{D}(t_0)$
    \begin{equation}\label{eq:s-step-initial-d-complexity}
      \mat{D}(t_0) = \mat{M}\mat{D}_0\mat{M}\transposed,
    \end{equation}
    \item Flow-field evaluation
    \begin{subequations}\label{eq:s-step-flowfield-complexity}
    \begin{align}
      \mat{U}_h\transposed F(t, \mat{U}_h\mat{D}(t)\mat{U}_h\transposed)\mat{U}_h
      &= \mat{U}_h\transposed \left( \driftmat \mat{U}_h\mat{D}(t)\mat{U}_h\transposed + \mat{U}_h\mat{D}(t)\mat{U}_h\transposed\driftmat\transposed + \dispmat\dispmat\transposed\right) \mat{U}_h\\
      &= \mat{U}_h\transposed(\driftmat(\mat{U}_h\mat{D}(t))) + (\mat{D}(t)\mat{U}_h\transposed)(\driftmat\transposed\mat{U}_h) + \mat{U}_h\transposed(\dispmat\dispmat\transposed\mat{U}_h).
    \end{align}
  \end{subequations}
  \end{enumerate}
  The above steps 1.\,and 2.\,are repeated according to how many DLRA integration steps are performed in the respective prediction step. Therefore, their cost has to be multiplied by that (typically small) constant.
  \item Using DLRA integration we obtain a low-rank factorization of the process-noise covariance matrix $\processnoisecov_l \approx \mat{Y}_l = \mat{U}_l\mat{D}_l\mat{U}_l^2$.
  It remains to compute a matrix square root $\processnoisecov_l\matsqrt = \mat{U}_l\mat{D}_l\matsqrt$ and thus
  a matrix square root of $\mat{D}_l \in \Re^{\lowrankdim \times \lowrankdim}$
  \begin{equation}\label{eq:post-dlra-sqrt-d}
    \mat{D}_l = \mat{D}_l\matsqrt\mat{D}_l\transposedsqrt.
  \end{equation}
\end{enumerate}

Now, having obtained $\processnoisecov\matsqrt_l$, we proceed to
\begin{enumerate}
  \setcounter{enumi}{3}
  \item predict the mean
  \begin{equation}\label{eq:predict-mean-complexity}
    \mu^-_{l} = \transitionmat_l\mu_{l-1}, 
  \end{equation}
  \item and the low-rank factor $\predcov\matsqrt_l$ of the predicted covariance matrix. This amounts to building the rank-$2\lowrankdim$ square-root factor and truncating
  it at its $\lowrankdim$-th largest singular value (i.e.\,, a truncated SVD of a tall $\statedim\times 2\lowrankdim$ matrix):
  \begin{equation}\label{eq:predict-cov-complexity}
    \begin{pmatrix}\transitionmat_l\lowrank{\filtcov}_{l-1}\matsqrt & \lowrank{\processnoisecov}\matsqrt_l\end{pmatrix} \approx \tilde{\mat{U}}_l\tilde{\mat{D}}_l\tilde{\mat{V}}_l\transposed.
  \end{equation}
\end{enumerate}
The computational complexities of the respective steps can be found in \cref{table:prediction-complexity}.

\begin{table}
  \caption{Time complexity of reduced-rank filtering: The correction step}
  \label{table:update-complexity}
  \centering
  \begin{tabular}{llll}
    \toprule
    \multicolumn{2}{c}{Step}\\
    \cmidrule(r){1-2}
    Eq. & Operation & Best case & Worst case\\
    \midrule
    \multirow{2}{*}{\eqref{eq:correction-svd-complexity}} & $\measurementnoisecov\matsqrtinv (\measurementmat \predcov_l\matsqrt)$ & $\mathcal{O}(\measdim\lowrankdim)$ & $\mathcal{O}(\measdim^2\lowrankdim)$ \\
    & $(\measurementnoisecov\matsqrtinv \measurementmat \lowrank{\predcov}_l\matsqrt )\transposed = \mat{U}_l \mat{D}_l \mat{V}_l\transposed$ & $\mathcal{O}(\measdim\lowrankdim^2)$ & $\mathcal{O}(\measdim\lowrankdim^2)$ \\
    \midrule
    \eqref{eq:correction-whitened-residual-complexity} & $\vec{e}_l = \measurementnoisecov\matsqrtinv(\vec{y}_l - \measurementmat \vec{\mu}_l^-)$ & $\mathcal{O}(\measdim)$ & $\mathcal{O}(\measdim\statedim + \measdim^2)$\\
    \midrule
    \eqref{eq:correction-mean-increment-complexity} & $\lowrank{\predcov}_l\matsqrt  (\mat{U}_l ((\mathrm{I} + \mat{D}_l^2)\matinv (\mat{D}_l (\mat{V}_l\transposed \vec{e}_l))))$ & $\mathcal{O}(\measdim\lowrankdim + \statedim\lowrankdim^2)$ & $\mathcal{O}(\measdim\lowrankdim + \statedim\lowrankdim^2)$\\
    \midrule
    \eqref{eq:correction-cov-update} & $\lowrank{\predcov}_l\matsqrt  (\mat{U}_l ( \mathrm{I} + \mat{D}_l^2)\matsqrtinv)$ & $\mathcal{O}(\statedim\lowrankdim^2)$ & $\mathcal{O}(\statedim\lowrankdim^2)$ \\
    \midrule
    \multirow{4}{*}{\eqref{eq:marginal-pred-loglik}}
    & $\log \abs[0]{\measurementnoisecov\matsqrt}$ & $\mathcal{O}(m)$ & $\mathcal{O}(m)$ \\[1mm]
    & $\sum_{k=1}^\lowrankdim \log ( (\mat{D}_l)_{kk}^2 + 1) $ & $\mathcal{O}(\lowrankdim)$ & $\mathcal{O}(\lowrankdim)$ \\[1mm]
    & $\norm{\vec{e}_l}^2$ & $\mathcal{O}(\lowrankdim)$ & $\mathcal{O}(\lowrankdim)$ \\
    & $(\vec{e}_l\transposed \mat{V}_l) (\mat{D}_l ((\mat{D}_l^2  + \mathrm{I})\matinv \mat{D}_l)) (\mat{V}_l\transposed \vec{e}_l)$ & $\mathcal{O}(\measdim\lowrankdim)$ & $\mathcal{O}(\measdim\lowrankdim)$ \\
    \bottomrule
  \end{tabular}
\end{table}

\paragraph{Correction step}
Now, the cost of the correction step is analyzed.
First, we compute an SVD of a wide $\lowrankdim\times\measdim$ matrix
\begin{equation}\label{eq:correction-svd-complexity}
  (\measurementnoisecov\matsqrtinv \measurementmat \lowrank{\predcov}_l\matsqrt )\transposed = \mat{U}_l \mat{D}_l \mat{V}_l\transposed. 
\end{equation}
Given this decomposition and the whitened residual
\begin{equation}\label{eq:correction-whitened-residual-complexity}
  \vec{e}_l = \measurementnoisecov\matsqrtinv(\vec{y}_l - \measurementmat \vec{\mu}_l^-), 
\end{equation}
we proceed to
\begin{enumerate}
    \item update the mean
      \begin{equation}\label{eq:correction-mean-increment-complexity}
        \Delta\mu_l = \mat{K}_l \vec{e}_l = \lowrank{\predcov}_l\matsqrt  (\mat{U}_l ((\mathrm{I} + \mat{D}_l^2)\matinv (\mat{D}_l (\mat{V}_l\transposed \vec{e}_l)))),
      \end{equation}
    \item and compute the low-rank factor of the filtering covariance
    \begin{equation}\label{eq:correction-cov-update}
      \lowrank{\filtcov}_l\matsqrt  = \lowrank{\predcov}_l\matsqrt  (\mat{U}_l ( \mathrm{I} + \mat{D}_l^2)\matsqrtinv). 
    \end{equation}
  \end{enumerate}

Finally, the marginal predictive log-likelihood $\log p(\vec{y}_l \mid \vec{y}_{1:l-1})$ is computed as in \cref{eq:marginal-pred-loglik}.

The computational complexities of the respective steps can be found in \cref{table:update-complexity}.
Together, \cref{table:prediction-complexity,table:update-complexity} prove the stated asymptotic complexities of the low-rank filtering recursion.
\end{proof}

\subsection{Proof of Proposition \ref{prop:smoothing-complexity}}\label{supp:subsec-proof-smoothing-complexity}

This section analyzes the computational complexity of the low-rank smoothing recursion in detail,
which proves \cref{prop:smoothing-complexity}.

\begin{proof} The best-case cost of the approximate smoothing scales linear in the state dimension $\statedim$.
  In the worst case, the complexity is quadratic in $\statedim$.

  \paragraph{Approximate backwards kernel}
  Factorizations of $\filtcov\matsqrt_{l-1}$ and $\predcov\matsqrt_l$ are already given from filtering.
  For the smoothing gain, it remains to compute
  \begin{equation}\label{eq:smoothing-gamma-complexity}
    \mat{\Gamma}_l =\lowrank{\filtcov}_{l}\transposedsqrt(\transitionmat_{l+1}\transposed(\lowrank{\predcov}_{l+1}\matsqrt)\matpseudoinv).
  \end{equation}
  Then, we proceed to compute the shift vector
  \begin{equation}\label{eq:smoothing-v-complexity}
    \vec{v}_l = \mu_l - \mat{G}_l\mu^-_{l+1} = \mu_l - \lowrank{\filtcov}_{l}\matsqrt(\mat{\Gamma}_l((\lowrank{\predcov}_{l+1}\transposedsqrt)\matpseudoinv\mu_{l+1}^-)), 
  \end{equation}
  and the low-rank factor $\mat{P}_l\matsqrt$ of the backwards-process-noise covariance matrix. Therefore, a truncated SVD of a tall $\statedim\times 2\lowrankdim$ matrix is computed:
  \begin{equation}\label{eq:smoothing-backwards-noise-cov-complexity}
    \begin{pmatrix}(\mat{I} - \mat{G}_l\transitionmat_{l+1}) \lowrank{\filtcov}_{l}\matsqrt & \mat{G}_l\lowrank{\processnoisecov}\matsqrt_{l+1}\end{pmatrix} \approx \widehat{\mat{U}}_l\widehat{\mat{D}}_l\widehat{\mat{V}}_l\transposed.
  \end{equation}

  \paragraph{Backwards prediction (smoothing)}
  Smoothing amounts to consecutive predictions with the backwards kernel.
  For this prediction, the low-rank factor of the process-noise covariance matrix does not come from DLRA, but can be directly
  computed from the smoothing gain and the process-noise covariance matrix (see \cref{eq:backwards-kernel-process-noise-cov-factorization,eq:backwards-kernel-process-noise-cov}).

  The smoothing mean is therefore given as
  \begin{equation}\label{eq:smoothing-mean-complexity}
    \xi_{l} = \mat{G}_l \xi_{l+1} + \vec{v}_l = \filtcov\matsqrt_{l}(\Gamma_l((\predcov_{l+1}\transposedsqrt)\matpseudoinv\xi_{l+1})) + \vec{v}_l,
  \end{equation}
  and the low-rank factor of the smoothing covariance as
  $\Lambda_{l}\matsqrt = \bar{\mat{U}}_l \bar{\mat{D}}_l$,
  where
  \begin{equation}\label{eq:smoothing-cov-complexity}
    \begin{pmatrix}
      \mat{G}_l\Lambda\matsqrt_{l+1} & \mat{P}\matsqrt_l
    \end{pmatrix}
    \approx \bar{\mat{U}}_l\bar{\mat{D}}_l\bar{\mat{V}}_l\transposed
  \end{equation}
  is the truncated SVD of a tall $\statedim \times 2\lowrankdim$ matrix.

  The computational complexities of the respective steps can be found in \cref{table:smoothing-complexity}, which shows
  that---in the best case---the cost of approximate low-rank smoothing never exceeds $\mathcal{O}(\statedim\lowrankdim^2 + \lowrankdim^3)$, as stated
  by \cref{prop:smoothing-complexity}. It also shows that in the worst case, the complexity is quadratic in $\statedim$.
\end{proof}

\begin{table}
  \caption{Time complexity of reduced-rank smoothing}
  \label{table:smoothing-complexity}
  \centering
  \begin{tabular}{llll}
    \toprule
    \multicolumn{2}{c}{Step}\\
    \cmidrule(r){1-2}
    Eq. & Operation & Best case & Worst case\\
    \midrule
    \eqref{eq:smoothing-gamma-complexity} & $\mat{\Gamma}_l =\lowrank{\filtcov}_{l-1}\transposedsqrt(\transitionmat_{l}\transposed(\lowrank{\predcov}_{l}\matsqrt)\matpseudoinv)$ & $\mathcal{O}(\statedim\lowrankdim^2)$ & $\mathcal{O}(\statedim^2\lowrankdim + \statedim\lowrankdim^2)$ \\
    \midrule
    \eqref{eq:smoothing-v-complexity} & $\mat{G}_l\mu^-_{l+1} = \lowrank{\filtcov}_{l}\matsqrt(\mat{\Gamma}_l((\lowrank{\predcov}_{l+1}\transposedsqrt)\matpseudoinv\mu_{l+1}^-))$ & $\mathcal{O}(\statedim\lowrankdim + \lowrankdim^2)$ & $\mathcal{O}(\statedim\lowrankdim + \lowrankdim^2)$\\
    \midrule
    \multirow{3}{*}{\eqref{eq:smoothing-backwards-noise-cov-complexity}} & $\mat{G}_l\transitionmat_{l+1}\filtcov\matsqrt_{l} = \filtcov\matsqrt_{l}(\Gamma_l((\predcov_{l+1}\transposedsqrt)\matpseudoinv(\transitionmat_{l+1}\filtcov_{l}\matsqrt)))$ & $\mathcal{O}(\statedim\lowrankdim^2 + \lowrankdim^3)$ & $\mathcal{O}(\statedim^2\lowrankdim + \statedim\lowrankdim^2 + \lowrankdim^3)$ \\
    & $\mat{G}_l\processnoisecov\matsqrt_l = \filtcov_{l-1}\matsqrt(\Gamma_l((\predcov_l\transposedsqrt)\matpseudoinv\processnoisecov\matsqrt_l))$ & $\mathcal{O}(\statedim\lowrankdim^2 + \lowrankdim^3)$ & $\mathcal{O}(\statedim\lowrankdim^2 + \lowrankdim^3)$ \\
    & $((\mat{I} - \mat{G}_l\transitionmat_{l+1}) \lowrank{\filtcov}_{l}\matsqrt \q \mat{G}_l\lowrank{\processnoisecov}\matsqrt_{l+1}) \approx \widehat{\mat{U}}_l\widehat{\mat{D}}_l\widehat{\mat{V}}_l\transposed$ & $\mathcal{O}(\statedim\lowrankdim^2)$ & $\mathcal{O}(\statedim\lowrankdim^2)$ \\
    \midrule
    \eqref{eq:smoothing-mean-complexity} & $\mat{G}_l \xi_l = \filtcov\matsqrt_{l-1}(\Gamma_l((\predcov_l\transposedsqrt)\matpseudoinv\xi_l))$ & $\mathcal{O}(\statedim\lowrankdim + \lowrankdim^2)$ & $\mathcal{O}(\statedim\lowrankdim + \lowrankdim^2)$ \\
    \midrule
    \multirow{2}{*}{\eqref{eq:smoothing-cov-complexity}} & $\mat{G}_l \Lambda_{l+1}\matsqrt = \filtcov\matsqrt_{l}(\Gamma_l((\predcov_{l+1}\transposedsqrt)\matpseudoinv\Lambda_{l+1}\matsqrt))$ & $\mathcal{O}(\statedim\lowrankdim^2 + \lowrankdim^3)$ & $\mathcal{O}(\statedim\lowrankdim^2 + \lowrankdim^3)$\\
    & $(\mat{G}_l\Lambda\matsqrt_{l+1} \q \mat{P}\matsqrt_l)
    \approx \bar{\mat{U}}_l\bar{\mat{D}}_l\bar{\mat{V}}_l\transposed$ & $\mathcal{O}(\statedim\lowrankdim^2)$ & $\mathcal{O}(\statedim\lowrankdim^2)$\\
    \bottomrule
  \end{tabular}
\end{table}

\section{Efficient inference for the case r > m}\label{supp:sec-r-larger-m}

This section gives an inference scheme that can be used in the correction step in case the low-rank dimension $\lowrankdim$ exceeds the measurement dimension $\measdim$ at a time point $t_l$.
This is not generally the designated use case of the algorithm, since in most applications, we assume $\lowrankdim \ll \measdim \leq \statedim$.
However, if at a given time point $t_l$, there are less than usual measurements available (e.g.\,due to missing data at that time), it is still useful to have
an inference scheme in place for this case. This is provided by the following proposition.

\begin{proposition}
  Let $\measdim < \lowrankdim \leq \statedim$ and $\predcov\matsqrt_l \in \Re^{\statedim \times \lowrankdim}$ and $\measurementnoisecov\matsqrt \in \Re^{\measdim \times \measdim}$.
  An approximate update is then obtained according to the standard Kalman filter update rules:
  \begin{subequations}
  \begin{align}
  \mat{S}_l &= \measurementmat \predcov_l\matsqrt (\measurementmat \predcov_l\matsqrt)\transposed + \measurementnoisecov\matsqrt\measurementnoisecov\transposedsqrt, \\
  \mat{K}_l &= \predcov_l\matsqrt (\measurementmat \predcov_l\matsqrt )\transposed \mat{S}\matpseudoinv, \\
  \Delta \vec{\mu}_l &= \mat{K}_l(\vec{y}_l - \measurementmat\vec{\mu}_l), \\
  \filtcov_l &= \predcov_l - \mat{K}_l \mat{S}_l \mat{K}_l\transposed.
  \end{align}
  \end{subequations}
  The marginal measurement covariance, $\mat{S}_l$, is obtained by a singular value decomposition
  \begin{subequations}\label{eq:r-larger-m-S-factorization}
  \begin{align}
  \mat{U}^s_l \mat{D}^s_l (\mat{V}^s_l)\transposed &= \begin{pmatrix} \measurementmat \predcov_l\matsqrt &  \measurementnoisecov\matsqrt \end{pmatrix}, \\
  \mat{S} &= \mat{U}^s_l (\mat{D}^s_l)^2 (\mat{U}^s_l)\transposed,
  \end{align}
  \end{subequations}
  where the matrix on the right-hand side of the first equation is $\measdim \times (\lowrankdim + \measdim)$.
  Hence, $\mat{U}^s_l \in \Re^{\measdim \times (\lowrankdim + \measdim)}$, and  $\mat{D}^s_l,\mat{V}^s_l \in \Re^{(\lowrankdim + \measdim) \times (\lowrankdim + \measdim)}$.\
  The Moore--Penrose pseudoinverse $\mat{S}_l\matpseudoinv$ is then given by
  \begin{equation}
  \mat{S}_l\matpseudoinv = \mat{U}^s_l (\mat{D}^s_l)^{-2} (\mat{U}^s_l)\transposed.
  \end{equation}
  The gain matrix is thus obtained as
  \begin{subequations}
  \begin{align}
  \widetilde{\mat{K}}_l   &= (\measurementmat \predcov_l\matsqrt )\transposed \mat{U}^s (\mat{D}^s)\matinv, \\
  \mat{K}_l &= \predcov_l\matsqrt \widetilde{\mat{K}}_l  (\mat{D}^s)\matinv (\mat{U}^s)\transposed,
  \end{align}
  \end{subequations}
  where $\widetilde{\mat{K}}_l \in \Re^{\lowrankdim \times (\lowrankdim + \measdim)}$.
  The updated covariance is obtained by
  \begin{equation}
  \begin{split}
  \filtcov_l &= \predcov_l - \mat{K}_l \mat{S}_l \mat{K}_l\transposed \\
  &=  \predcov_l - \predcov_l\matsqrt \widetilde{\mat{K}}_l (\mat{D}_l^s)\matinv (\mat{U}_l^s)\transposed \mat{U}_l^s \mat{D}_l^s (\mat{U}_l^s)\transposed ( \predcov_l\matsqrt \widetilde{\mat{K}}_l (\mat{D}_l^s)\matinv (\mat{U}_l^s)\transposed )\transposed \\
  &=  \predcov_l - \predcov_l\matsqrt \widetilde{\mat{K}}_l ( \predcov_l\matsqrt \widetilde{\mat{K}}_l )\transposed \\
  &=  \predcov_l\matsqrt\big( \mathrm{I} - \widetilde{\mat{K}}_l \widetilde{\mat{K}}_l\transposed \big) \predcov_l\transposedsqrt.
  \end{split}
  \end{equation}
  Consider the singular value decomposition of $\widetilde{\mat{K}}_l$
  \begin{equation}
  \widetilde{\mat{K}}_l = \mat{U}_l^k \mat{D}_l^k (\mat{V}^k_l)\transposed,
  \end{equation}
  where $\mat{U}_l^k,\mat{D}_l^k  \in \Re^{\lowrankdim\times \lowrankdim}$ and $\mat{V}_l^k \in \Re^{\lowrankdim \times \lowrankdim + \measdim}$.
  The covariance update in low-rank form is then given by
  \begin{equation}
  \begin{split}
  \Sigma\matsqrt &= \predcov\matsqrt\big(\mathrm{I} -   \mat{U}_l^k (\mat{D}_l^k)^2 (\mat{U}_l^k)\transposed \big)\matsqrt, \\
  &=  \predcov\matsqrt \big(\mat{U}_l^k (\mat{U}_l^k)\transposed -   \mat{U}_l^k (\mat{D}_l^k)^2 (\mat{U}_l^k)\transposed \big)\matsqrt, \\
  &= \predcov\matsqrt \mat{U}_l^k \big( \mathrm{I} -    (\mat{D}_l^k)^2 \big)\matsqrt.
  \end{split}
  \end{equation}
  The marginal predictive log likelihood $p(\vec{y}_l \mid \vec{y}_{1:l-1}) = \N(\measurementmat\mu^-_l, \mat{S}_l)$ can be cheaply evaluated given the factorization of $\mat{S}_l$ from \cref{eq:r-larger-m-S-factorization}.

  \end{proposition}

\section{Dynamical-low-rank approximation algorithm for Lyapunov equations}\label{supp:sec-dlra}

Following \citet{Ceruti2022}, we give the procedure for one DLRA integration step, adapted to our specific case of
a symmetric Lyapunov equation
\begin{equation}
  F(t, \processnoisecov(t)) = \driftmat\processnoisecov(t) + \processnoisecov(t)\driftmat\transposed + \dispmat(t)\dispmat(t)\transposed.
\end{equation}
Let an initial rank-$\lowrankdim$ factorization $\processnoisecov_0 \approx \mat{Y}_0 = \mat{U}_0\mat{D}_0\mat{U}_0\transposed$, with an orthogonal matrix $\mat{U}_0 \in \Re^{\statedim \times \lowrankdim}$ and a symmetric matrix $\mat{D}_0 \in \Re^{\lowrankdim \times \lowrankdim}$, be given at time $t_0$.
For a temporal step size $h$, a rank-$\lowrankdim$ factorization $\processnoisecov(t + h) \approx \mat{Y}_h = \mat{U}_h\mat{D}_h\mat{U}_h\transposed$ at the next integration step $t_0 + h$
is computed as follows.
\begin{enumerate}
  \item[] \textbf{K-step:} Update $\mat{U}_0 \rightarrow \mat{U}_h$. \\
  Integrate from $t = t_0$ to $t_0 + h$ the $\statedim \times \lowrankdim$ matrix differential equation
  \begin{equation}\label{eq:dlra-k-step}
    \dot{\mat{K}}(t) = F(t, \mat{K}(t)\mat{U}_0\transposed)\mat{U}_0, \qq \mat{K}(t_0) = \mat{U}_0\mat{D}_0.
  \end{equation}
  Orthogonalize $K(t+h)$ by computing a QR decomposition, yielding the orthogonal matrix $\mat{U}_h$. Then, compute the $\lowrankdim \times \lowrankdim$ matrix $\mat{M} = \mat{U}_h\transposed \mat{U}_0$.
  \item[] \textbf{S-step:} Update $\mat{D}_0 \rightarrow \mat{D}_h$. \\
  Integrate from $t = t_0$ to $t_0 + h$ the $\lowrankdim \times \lowrankdim$ matrix differential equation
  \begin{equation}\label{eq:dlra-s-step}
    \dot{\mat{D}}(t) = \mat{U}_h\transposed F(t, \mat{U}_h\mat{D}(t)\mat{U}_h\transposed)\mat{U}_h, \qq \mat{D}(t_0) = \mat{M}\mat{D}_0\mat{M}\transposed,
  \end{equation}
  and set $\mat{D}_h = \mat{D}(t + h)$
\end{enumerate}
\begin{remark}
  The terms ``K-step'' and ``S-step'' can be confusing due to conflicting notation conventions in the Kalman-filtering and dynamical-low-rank-approximation literature, respectively.
  The matrix $\mat{K}$ and the term ``S-step'' are not related to the Kalman gain $\mat{K}$ and the marginal measurement covariance matrix $\mat{S}$ from the Kalman filter update step described in \cref{subsec:filtering,supp:sec-r-larger-m}.
\end{remark}
The $\statedim \times \lowrankdim$ and $\lowrankdim \times \lowrankdim$ matrix differential equations in the K-step and S-step can be solved
using a standard numerical integrator, e.g., a Runge-Kutta method or an exponential integrator.

It is useful to note that the matrix equation in the S-step itself is a symmetric Lyapunov equation:
\begin{subequations}
  \begin{align}
    \dot{\mat{D}}(t) &= \mat{U}_h\transposed F(t, \mat{U}_h\mat{D}(t)\mat{U}_h\transposed)\mat{U}_h\\
    &= \mat{U}_h\transposed \left( \driftmat\mat{U}_h\mat{D}(t)\mat{U}_h\transposed + \mat{U}_h\mat{D}(t)\mat{U}_h\transposed\driftmat\transposed + \dispmat\dispmat\transposed\right) \mat{U}_h\\
    &= (\mat{U}_h\transposed\driftmat\mat{U}_h)\mat{D}(t) + \mat{D}(t)(\mat{U}_h\transposed\driftmat\mat{U}_h)\transposed + \mat{U}_h\transposed\dispmat\dispmat\transposed\mat{U}_h \\
    &= \driftmat_\mat{D}\mat{D}(t) + \mat{D}(t)\driftmat\transposed_\mat{D} + \dispmat_\mat{D}\dispmat_\mat{D}\transposed,
  \end{align}
\end{subequations}
where the final step defines the parameters of the Lyapunov equation for $\mat{D}(t)$ as $\driftmat_\mat{D} = \mat{U}_h\transposed\driftmat\mat{U}_h$ and $\dispmat_\mat{D} = \mat{U}_h\transposed\dispmat$.
Since $\mat{D}(t) \in \Re^{\lowrankdim \times \lowrankdim}$ is small,
this equation can be solved exactly with little computational cost using matrix-fraction decomposition \citep{Stengel1994,Axelsson2015}.

\section{Details on experimental setups}\label{supp:sec-exp-details}
This section gives more details regarding the experimental setups used in \cref{sec:experiments}.
For all results of the proposed RRKF algorithm, only a single DLRA step was used in the prediction step to compute the low-rank factorization of the process noise covariance matrix $\processnoisecov\matsqrt$.
\subsection{Linear advection model}\label{supp:subsec-la-exp}
This section provides more details on the data assimilation setting that is considered in \cref{subsec:LA-exp}.

We consider a spatial grid of $\statedim = 1024$ uniformly spaced points and a temporal grid of 800 uniformly spaced time points.
Unit step sizes $\Delta t = \Delta x = 1$ are assumed on the spatio-temporal grid.
To generate an initial ground-truth state we sample an initial sinusoidal curve $\psi(0)$ according to the description by \citet{Sakov2008}
\begin{equation}\label{eq:sinusoidal-LA}
  \psi_i(0) = \sum_{k = 0}^{25} a_k \sin\left(\frac{2 \pi k}{1000} i + \varphi_k\right),
\end{equation}
where $i = 1, \dots, 1024$ is an index into the spatial grid and $a_k \sim \operatorname{Unif}(0, 1)$, $\varphi_k \sim \operatorname{Unif}(0, 2\pi)$.
The initial state is normalized as described by \citet{Sakov2008}.
To generate a ground-truth trajectory from this initial state, the linear-advection dynamics $\frac{\partial \psi}{\partial t} = -\alpha \frac{\partial \psi}{\partial x}$ are simulated on
the finite spatial grid. We assume constant unit velocity $\alpha = 1$ and periodic boundary conditions in space.
As data, 10 equidistant state components are observed every 5 time steps.
Each observation is corrupted by additive Gaussian noise.

An initial ensemble of size $\lowrankdim = \statedim$ is built by successive sampling according to \cref{eq:sinusoidal-LA}.
The exact sampling process follows the more detailed description by \citet[Section 3.1]{Evensen2021}.
From this $\Re^{\statedim \times \statedim}$ ensemble matrix, the sample covariance matrix is computed and used as the initial covariance for the Kalman filter.
The initial factorization of the RRKF is obtained by a spectral decomposition of the sample covariance matrix truncated at the $\lowrankdim$-th largest eigenvalue.
Selecting the $\lowrankdim$ first ensembles serves as an initial ensemble for the EnKF and the ETKF.

\Cref{fig:LA-subplot} shows the deviations of the individual low-rank approximations to the optimal KF estimate on the described data assimilation problem.

\subsection{London air-quality regression}

The experimental setup and the data used in \cref{subsec:london-exp} is provided by \citet{Hamelijnck2021}.
The model is selected via the log-likelihood estimation of the \emph{exact} Kalman filter.
The RMSE of the Kalman filter mean to the test data is $9.96791$ (cf.\,\citet{Hamelijnck2021}).
For $\lowrankdim = \statedim$ the RRKF also obtains this RMSE up to numerical error, as shown in \cref{fig:london-subplot}.

\subsection{Spatio-temporal Mat\'ern process with varying spatial lengthscale}\label{supp:subsec-on-model-exp}

\Cref{subsec:on-model-exp} evaluates the low-rank approximation quality in
approximate on-model spatio-temporal GP regression with varying strength of interaction between the state components.
In this experiment, we consider a spatio-temporal GP with separable covariance structure, as described in \cref{subsec:on-model-exp} with hand-picked hyperparameters.
The time domain is a uniform grid on the interval $[0.1, 10]$ with step size $\Delta t = 0.1$.
The spatial domain is a uniform grid on $[0, 2] \times [0, 2]$ with step size $\Delta x = 0.1$.

Over the experiment, the spatial lengthscale is varied in order to evaluate low-rank approximation quality given how much correlation
between the respective state components is encoded in the prior.
The smaller the spatial lengthscale, the less interactions between the spatial point and the worse the low-rank approximation is expected to be for small $\lowrankdim$.
We test the values $\ell_x \in \{0.01, 0.1, 0.25, 1.0\}$.
For each of those values, we proceed as follows.
First, a realization of the prior dynamics is drawn from the spatio-temporal Mat\'ern process.
At every temporal grid point, the entire state vector of the prior draw at that point is corrupted by additive Gaussian noise to generate data.
Then, the Kalman filter estimate and the respective low-rank filtering estimates are computed for increasing values of $\lowrankdim$.
\Cref{fig:different_lx_r_vs_error} shows for increasing spatial lengthscales and varying $\lowrankdim$ (i) the resulting RMSE of the approximate filter means to the Kalman filter mean and (ii) the time-averaged Frobenius distance between the approximate filter covariance matrices to the Kalman filter covariance matrix.

\subsection{Runtime}
This section details the respective experimental setups used to evaluate the best-case and worst-case computational complexity of the approximate low-rank filters
with respect to the state dimension $\statedim$, as described in \cref{subsec:runtime-exp}.
We analyze the asymptotic computational complexity with respect to the state dimension $\statedim$
both for the linear-in-$\statedim$ case (\cref{fig:time_plot}, left) and for the quadratic-in-$\statedim$ case (\cref{fig:time_plot}, right).
The elapsed wall time is always measured using the BenchmarkTools.jl software package\footnote{\url{https://github.com/JuliaCI/BenchmarkTools.jl}} with default settings, which computes sample statistics over multiple runs
automatically to prevent distortions by background processes.
The setups for both analyzed cases differ only marginally from already encountered setups.

\paragraph{Linear-in-$\statedim$ case} We use a setting that is similar to what is described in \cref{subsec:LA-exp,supp:subsec-la-exp}.
The state dimension is varied in order to evaluate computational complexity with respect to $\statedim$.
The measurement dimension is fixed at $\measdim = 100$ and the low-rank dimension at $\lowrankdim = 5$.
The temporal domain is a grid on the interval $[0, 100]$ with unit step size $\Delta t = 1$.
Noisy observations are generated from the ground truth trajectory at every $5$ time points.

The discretized linear advection dynamics with periodic boundary conditions amount to multiplication with a circulant matrix.
The shift direction depends on the advection velocity, which we chose as $\alpha = 1$.
Multiplication with a circulant matrix can be carried out in linear complexity via multiplication in the Fourier space.
Since there is no process noise assumed, the matrix $\dispmat\dispmat\transposed$ is the zero matrix.
Hence, \cref{assumption:cheap-prediction} is satisfied by the dynamics model.

The measurement operator selects $\measdim = 100$ uniformly spaced spatial points from the state vector and
the measurement noise covariance matrix is a diagonal matrix.
Hence, the measurement model fulfills \cref{assumption:cheap-observation,assumption:diagonal-R}.

\paragraph{Quadratic-in-$\statedim$ case} Here, we use a setting that is similar to what is described in \cref{subsec:on-model-exp,supp:subsec-on-model-exp}.
The measurement dimension is fixed at $\measdim = 100$ and the low-rank dimension at $\lowrankdim = 5$.
The temporal domain is a grid on the interval $[0, 10]$ with step size $\Delta t = 0.1$.
The spatial domain are $\statedim$ equidistant points in an interval in $\Re$, where $\statedim$ varies.
The generated data are noisy observations from a ground-truth realization of the prior at $20$ time points, each measuring $\measdim = 100$ state components.
We evaluate the runtime of solving a spatio-temporal regression problem with this prior and the generated data.

\Cref{assumption:cheap-prediction} is \emph{not} fulfilled by the dynamics model in this setting, in that evaluating the map $x\mapsto \dispmat\dispmat\transposed x$ costs $\mathcal{O}(\statedim^2)$.
By modeling spatial diffusion with a dense spatial kernel matrix, the inhomogeneity in the Lyapunov equation for the process-noise covariance matrix is also a dense matrix.
Concretely, $\dispmat\dispmat\transposed = (\widetilde{\dispmat}\mathcal{K}_x\matsqrt)(\widetilde{\dispmat}\mathcal{K}_x\matsqrt)\transposed$, where $\widetilde{\dispmat}$ is the temporal dispersion matrix and
$\mathcal{K}_x$ denotes the spatial kernel matrix.
Since $\mathcal{K}_x$ is dense, so is $\dispmat\dispmat\transposed$, which violates a part of \cref{assumption:cheap-prediction}.
The cost of the low-rank prediction step thus scales quadratically in the state dimension $\statedim$.

The observation model is a sparse selection operator that projects the state onto $m = 100$ points and the measurement noise covariance is a diagonal matrix.
Hence, \cref{assumption:cheap-observation,assumption:diagonal-R} are still fulfilled by the observation model.

\Cref{fig:time_plot} demonstrates that the theoretical complexities are empirically verified.

\subsection{Large-scale spatio-temporal GP regression on rainfall data}
This section provides further details on the large-scale low-rank approximation to spatio-temporal GP regression, described in \cref{subsec:australia-exp}.

We select the prior model via the log-likelihood estimate of the RRKF, as given in \cref{eq:marginal-pred-loglik}.
For the model selection, the data set is distinct from the data used in the filtering/smoothing problem (\cref{fig:australia_rainfall}).
We use the time period from 25 January, 2010 through 5 March, 2010
and each data point is downsampled to a lower spatial resolution using cubic-spline interpolation.
This reduces the state dimension to $\statedim = 4350$ spatial points during model selection.
Further, for model selection the low-rank dimension is set to $\lowrankdim = 200$.

\section{Z-scores of (approximate) Gaussian state estimates}\label{supp:sec-calibration}
This sections investigates a limitation of the proposed algorithm, which is addressed in \cref{sec:limitations}.
When truncating covariance information between the state components, we expect this to be reflected by a \emph{higher} uncertainty in the resulting estimate.
However, the algorithm does not account for that missing information and tends to return overconfident estimates for small values of $\lowrankdim$.

We analyze the distribution of the Z-scores of Gaussian state estimates on a Gaussian model.
We expect the absolute values of the Z-scores to be distributed according to a Chi(1) distribution.
A spatio-temporal GP regression problem in an on-model setting, similar to the setting described in \cref{supp:subsec-on-model-exp}, is solved to investigate this.
A ground truth realization is drawn from a spatio-temporal Mat\'ern-\nicefrac{1}{2} process in the temporal domain $[0, 50]$ with a step size of $\Delta t = 0.1$.
Data is generated by adding zero-mean Gaussian noise to $\measdim = 150$ state components at $100$ randomly sampled time points.
The spatial grid is a uniformly spaced grid in the interval $[0, 20]$ with spatial step size of $\Delta x = 0.1$.

\begin{figure}
  \centering
  \includegraphics[width=.98\linewidth]{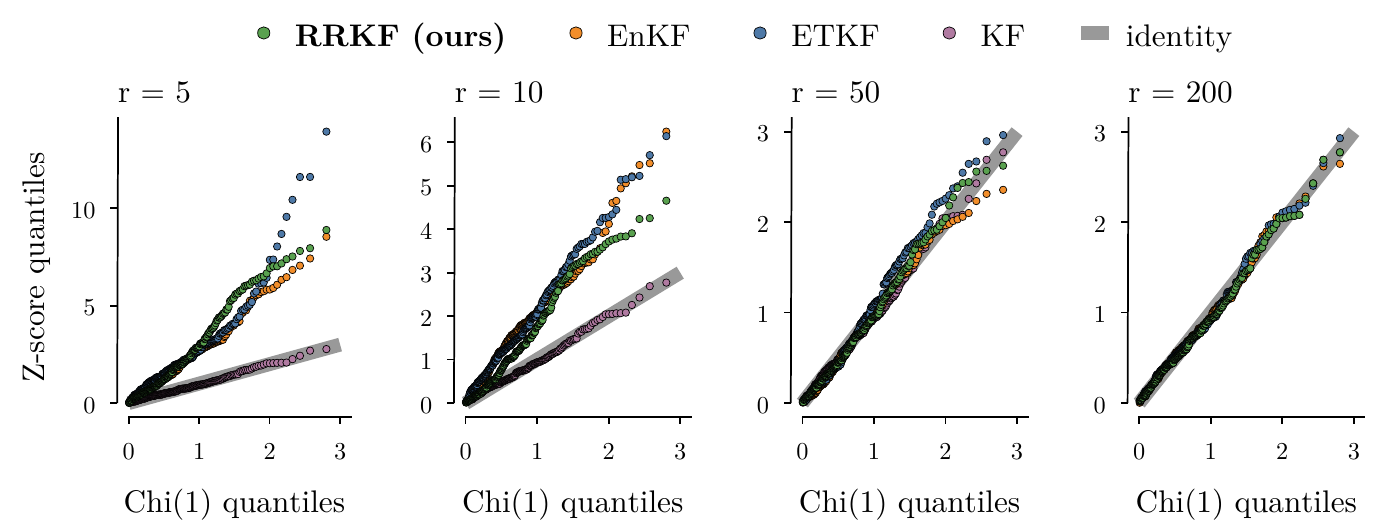}
  \caption{\emph{Z-score distribution for the low-rank filters with varying low-rank dimensions $\lowrankdim$.}
  In this on-model GP regression problem, we assume the Z-scores to be Chi(1) distributed.
  For the approximate low-rank filters, the Z-score distribution has too much mass in the high regimes, indicating
  overconfident estimates. For $\lowrankdim = \statedim$, the Z-scores of the RRKF and the KF align.
  }
  \label{fig:z-score-distribution}
\end{figure}

\Cref{fig:z-score-distribution} visualizes the distribution of the vector of Z-scores $\frac{\mu_\numT - \vec{x}^\star_\numT}{\sigma_\numT}$, where
$\mu_\numT$ and $\sigma_\numT$ are the final-step filtering/smoothing mean and standard deviation, respectively. $\vec{x}^\star_\numT$ is the ground-truth state at time $t_\numT$.
The Z-scores are computed for increasing values of $\lowrankdim$. The first $\lowrankdim = 50$ eigenvalues of the final-step Kalman filter covariance matrix account for around $97.8\%$ of the spectrum.
No covariance inflation \citep[Section 4.4]{Carrassi2018} is used for the ensemble methods.

It becomes apparent that---especially for very small values of $\lowrankdim$---there are significantly more high Z-scores than expected.
This indicates that too many states are estimated poorly and divided by a small standard deviation.
In their na\"ive implementation, all examined low-rank algorithms exhibit this behavior.
For the RRKF, it is left for future work to find a principled solution to account for missing covariance information in a computationally efficient manner.

\section{Error as a Function of Raw Computation Time}\label{supp:sec-err-vs-runtime}
\begin{figure}
  \centering
  \begin{subfigure}[b]{0.45\linewidth}
      \centering
      \includegraphics[width=\linewidth]{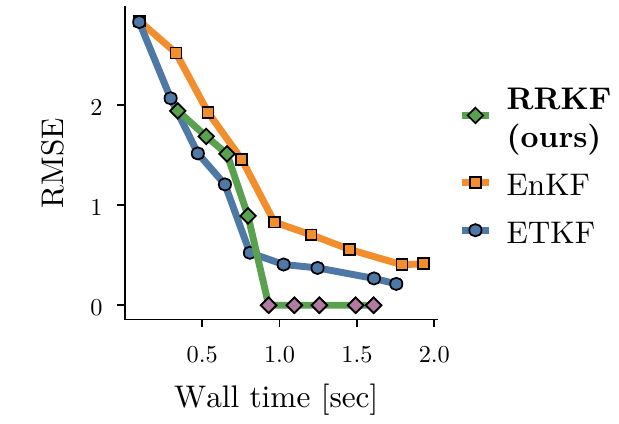}
      \caption{\emph{Linear advection dynamics (true rank = 51).}}
      \label{fig:LA-err-runtime}
  \end{subfigure}
  \hfill\\[4mm]
  \begin{subfigure}[b]{\linewidth}
      \centering
      \includegraphics[width=\linewidth]{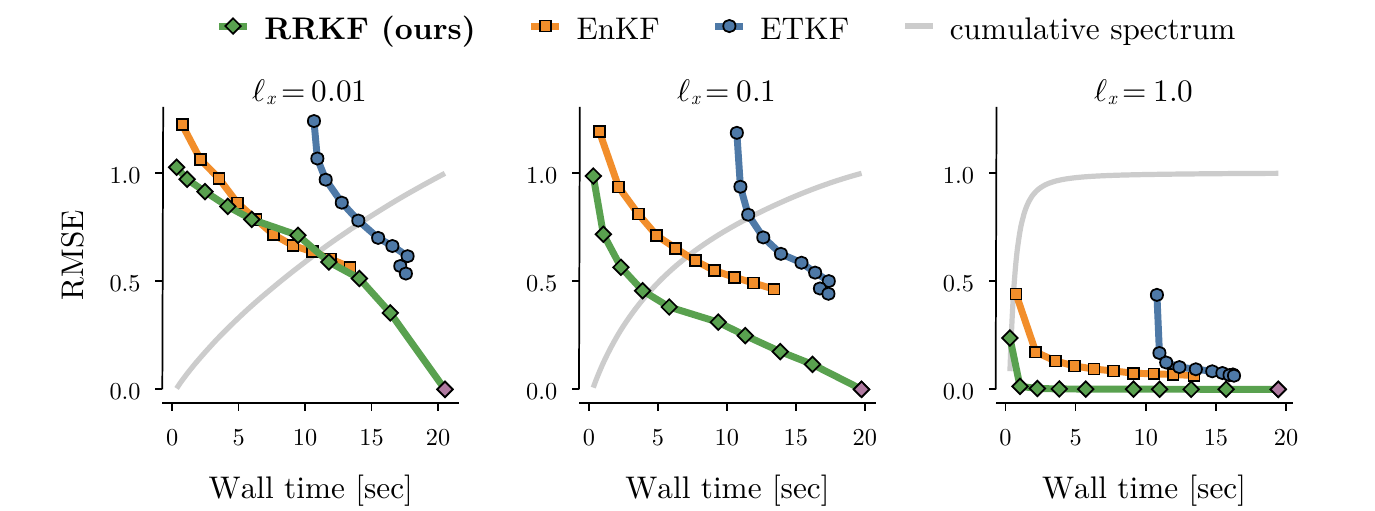}
      \caption{\emph{London air-quality regression.}}
      \label{fig:onmodel-err-runtime}
  \end{subfigure}
  \caption{\emph{Error of the low-rank filters as a function of wall-clock computation time.}
  The raw computational expenses are comparable to those of the ensemble methods. The lower-left corner is the optimal setting with low error and low computation time.
  All methods approach this region of the plot for faster decays of the spectrum, while the RRKF performs better most of the time.}
  \label{fig:err-runtime}
\end{figure}

In addition to the previous anaylsis of the error in the approximate filtering estimate with varying low-rank dimensions,
\Cref{fig:LA-err-runtime,fig:onmodel-err-runtime} show the error as a function of wall-clock computation times for the linear-advection model (\cref{subsec:LA-exp}) and the
spatio-temporal Mat\'ern model (\cref{subsec:on-model-exp}), respectively.

